%% file: main.tex
\documentclass[preprint,review,12pt]{elsarticle}\usepackage{hyperref}
\hypersetup{
colorlinks=true,
urlcolor=blue,
citecolor=blue}
\usepackage{amsfonts,amssymb,amsmath,amsgen,amsopn,amsbsy,theorem,graphicx,epsfig}
\usepackage{amscd,latexsym,mathrsfs,eurosym,enumerate}
\usepackage[utf8]{inputenc}
\usepackage[english]{babel}
\usepackage{cleveref,multirow}
\usepackage[dvipsnames]{xcolor}
\usepackage[pagewise]{lineno}
\graphicspath{{figures/}}

\journal{Robotics and Autonomous Systems}

\begin{document}

\newcommand\blfootnote[1]{%
  \begingroup
  \renewcommand\thefootnote{}\footnote{#1}%
  \addtocounter{footnote}{-1}%
  \endgroup
}
\begin{frontmatter}
\title{Object and Relation Centric Representations for Push Effect Prediction}

\author{
Ahmet E. Tekden$^{1,2}$, Aykut Erdem$^{3}$, Erkut Erdem$^{4}$, Tamim Asfour$^{5}$, and Emre Ugur$^1$}

\address{$^{1}$Computer Engineering Department, Bogazici University, Turkey }
\address{$^{2}$Electrical Engineering Department, Chalmers University of Technology, Sweden }
\address{$^{3}$Computer Engineering Department, Ko\c{c} University, Turkey}
\address{$^{4}$Computer Engineering Department, Hacettepe University, Turkey}
\address{$^{5}$Institute for Anthropomatics and Robotics, Karlsruhe Institute of Technology, Germany\\}
\ead{tekdenahmet@gmail.com}



\setcounter{page}{1}


\begin{abstract}
Pushing is an essential non-prehensile manipulation skill used for tasks ranging from pre-grasp manipulation to scene rearrangement, reasoning about object relations in the scene, and thus pushing actions have been widely studied in robotics. The effective use of pushing actions often requires an understanding of the dynamics of the manipulated objects and adaptation to the discrepancies between prediction and reality. For this reason, effect prediction and parameter estimation with pushing actions have been heavily investigated in the literature. However, current approaches are limited because they either model systems with a fixed number of objects or use image-based representations whose outputs are not very interpretable and quickly accumulate errors. In this paper, we propose a graph neural network based framework for effect prediction and parameter estimation of pushing actions by modeling object relations based on contacts or articulations.  Our framework is validated both in real and simulated environments containing different shaped multi-part objects connected via different types of joints and objects with different masses, and it outperforms image-based representations on physics prediction. Our approach  enables the robot to predict and adapt the effect of a pushing action as it observes the scene. It can also be used for tool manipulation with never-seen tools. Further, we demonstrate 6D effect prediction in the lever-up action in the context of robot-based hard-disk disassembly.
\end{abstract}
\begin{keyword}
Push Manipulation \sep Effect Prediction \sep Parameter Estimation \sep Graph Neural Networks \sep  Interactive Perception \sep Articulation Prediction
\end{keyword}
\end{frontmatter}
\input{sections/intro}
\input{sections/related_work}
\input{sections/method}
\input{sections/experiments}
\input{sections/conclusion}
\input{sections/acknowledgment}

\bibliographystyle{elsarticle-num}
\bibliography{references}

\end{document}

%% file: sections/intro.tex
\section{Introduction}
\label{sec:introduction}

\begin{figure}[!t]
    \centering
    \includegraphics[width=\linewidth]{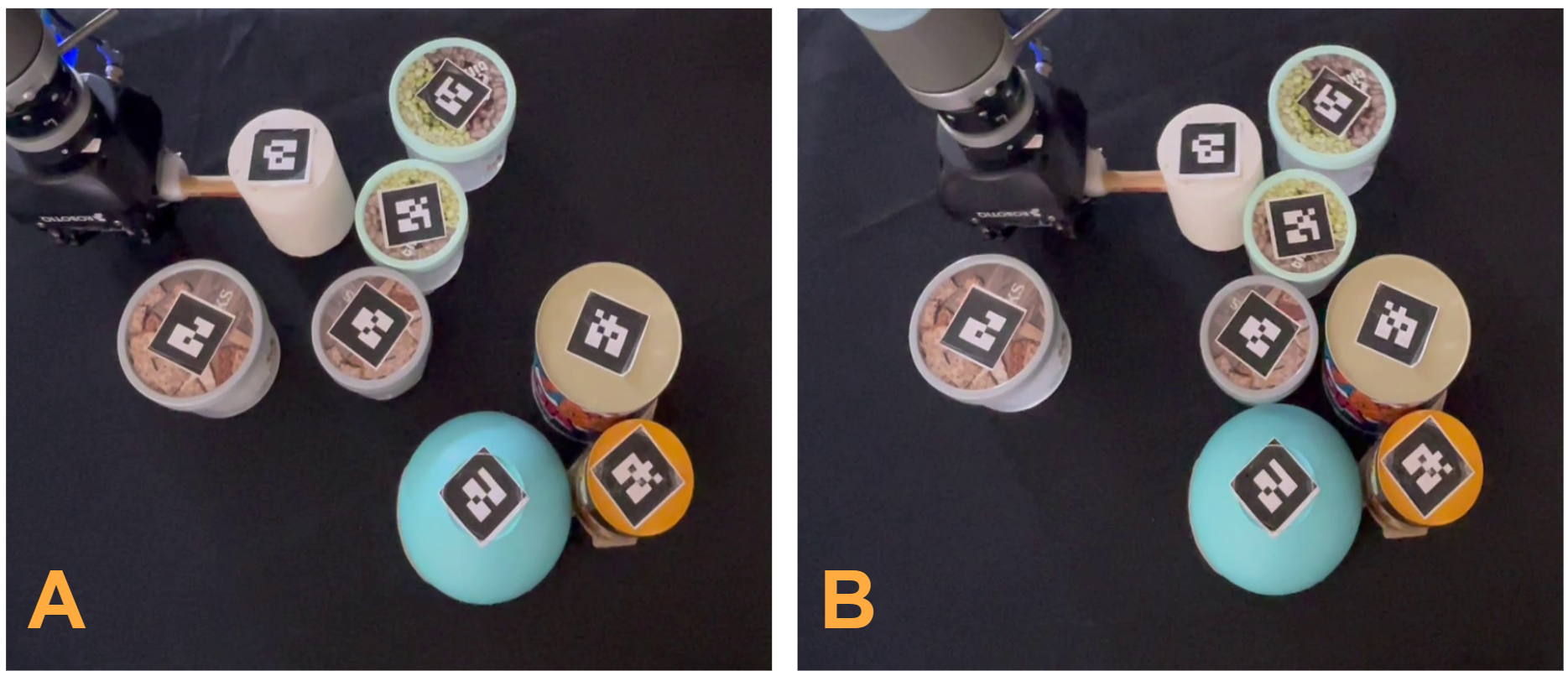}
    \caption{We will normally expect the action of the robot on the left image to scatter contacted objects. However, seeing the contacted objects moving together, the robot should correct its belief to enable this dynamic.}
    \label{fig:first_page}
\end{figure}

Pushing is a fundamental non-prehensile (manipulation without grasping) motion primitive that gives robots great flexibility in manipulating objects \cite{ruggiero2018nonprehensile,stuber2020let}. Using push actions, a robot can navigate objects to goal configurations even when objects are not graspable \cite{stuber2018feature}; it can manipulate objects under uncertainty \cite{dogar2010push}, or bring an object to the graspable area\cite{king2013pregrasp}. Compared to grasping actions, it is not as restrictive; however, the issue is that the robot does not have direct control over the state of the manipulated objects. This results in greater complexity in planning and control as the dynamics of the manipulated objects are often required to be taken into consideration\cite{ruggiero2018nonprehensile}. Effect prediction of pushing action has many applications\cite{stuber2020let,paus2020predicting}, including scene rearrangement\cite{mericcli2015push}, object segmentation \cite{van2012maximally}, object singulation\cite{eitel2020learning, zeng2018learning}, pre-grasp manipulation \cite{zeng2018learning, omrvcen2009autonomous, kappler2012templates,elliott2016making}. However, action-effect prediction of pushing actions depends on many factors\cite{yu2016more} and requires adaptation when mispredictions occurs. Figure~\ref{fig:first_page} shows an example illustration.  The initial prediction of the robot will be objects getting scattered. However, after seeing some of the objects moving together, the robot will understand that their future motion will continue reflecting this dynamic.  

In many environments, robots work with object clutters containing different shaped and weighted objects with possible articulations between them. A robot should be able to reason about the influence of shape and mass of objects, physical connections like contacts or different types of articulations between objects, propagation of motions between objects, and correction of unknown or partially known objects or object parts in the environment. Current approaches model environments with a fixed number of objects or use image data, an object-independent representation. While there has been great progress on effect prediction using raw sensory data\cite{finn2016unsupervised, 7989324, byravan2017se3, nematollahi2020hindsight}, using them on decision making level has been difficult and required tasks to be generated on pixel level. While there are certain advantages of such approaches, many tasks often require more interpretable representations for the task to be defined. Humans decompose environments into objects and use their interactions for physical reasoning~\cite{spelke1992origins, tenenbaum2011grow, mrowca2018flexible}, so there is certainly value in using such representations in effect prediction. We propose using graph neural networks (GNNs) for push effect prediction. Graph neural networks\cite{47094} can exploit the graph structure of multi-objects systems by exploiting and using object- and relation-centric representations and they are heavily used in modelling physics~\cite{battaglia2016interaction,chang2016compositional,li2018propagation,mrowca2018flexible,li2018learning,watters2017visual,van2018relational,sanchez2020learning}.

In this paper, we propose a general-purpose learnable physics engine in which object- and relation-centric representations are learned via a shared propagation network and used for physics prediction and parameter estimation in push manipulation tasks\footnote{Project page: \url{https://fzaero.github.io/push_learning/}}. We use articulation based graph representations that use cylinder- and cuboid-shaped objects and their possible interactions via contacts or joints for modeling multi-part object systems. We resort to a two-step training scheme where our framework is first trained for effect prediction, then using learned object and relation representations, it is trained for parameter estimation. Our framework can predict low-level trajectories of groups of articulated objects given robot actions and estimate the mass of observed objects and joint relations between them based on their interaction history. Using articulation based representation, novel tools that are not encountered during training can be built by connecting multiple cuboids via fixed joints, and they can be used in planning in tool manipulation tasks.
In addition,  we have shown that our framework can make 6D effect predictions. 
Compared to previous work~\cite{tekden2019belief}, our method acquires lower prediction errors for long-horizon prediction tasks (in Section~\ref{sec:real_world}). More specifically, the general contributions of our framework can be listed as follows: 
\begin{itemize}
    \item We develop a graph neural network based framework for parameter estimation and physics prediction in push manipulation tasks.
    \item We utilize a weight-sharing mechanism to transfer learned representations to be used in new tasks.
    \item We show the feasibility of articulation based graph representations for modeling multi-part objects and show that it outperforms object-centric image based representation in physics prediction task.
    \item We design a novel 6-D action-effect prediction in lever-up task in the context of  hard-disk drive disassembly.
    \item Through simulated and real-world experiments, we verify our framework in joint relation and mass prediction, physics prediction, and tool manipulation and planning tasks.
\end{itemize}

%% file: sections/related_work.tex
\section{Related Work}

\paragraph{Learning Dynamics / Modelling Physics}
Modeling intuitive physics has attracted considerable interest in recent years \cite{kubricht2017intuitive}. For instance, Battaglia \textit{et al.} \cite{battaglia2013simulation} proposed a Bayesian model called Intuitive Physics Engine and showed that the physics of stacked cuboids could be modeled with this model. Similarly, Hamrick \textit{et al.} \cite{hamrick2016inferring} showed that humans could reason about object masses from their interactions and modeled it with Bayesian models. Smith \textit{et al.} \cite{smith2019modeling} have modelled expectation violation in intuitive physics. They discuss how humans surprise when their physical expectations mismatch with reality, and they modeled this with deep learning methods. Deisenroth \textit{et al.} \cite{deisenroth2011pilco} suggested a probabilistic dynamic model that depends on Gaussian Processes and that is capable of predicting the next state of a robot given the current state and the action. Recently, these studies have been extended through the use of deep learning methods. Lerer \textit{et al.}~\cite{lerer2016learning} trained a deep network to predict the stability of the block towers given their raw images obtained from a simulator. Groth \textit{et al.} \cite{groth2018shapestacks} extended this idea by allowing stacking of objects with different geometries. They showed that their proposed network could predict the stability of given towers in this more difficult setup. The tower stacking task has continued to be an important environment for intuitive physics problems\cite{li2016fall}.

A specific topic of interest within modeling physics with deep learning is motion prediction from images, which has gained increasing attention over the last few years.  Mottaghi \textit{et al.} \cite{mottaghi2016newtonian} trained a Convolutional Neural Networks (CNN) for motion prediction on static images by casting this problem as a classification problem. Mottaghi \textit{et al.} \cite{mottaghi2016happens} employed CNNs to predict movements of objects in static images in response to applied external forces. Fragkiadaki \textit{et al.} \cite{fragkiadaki2015learning} suggested a deep architecture in which the outputs of a CNN are used as inputs to Long Short Term Memory (LSTM) cells~\cite{hochreiter1997long} to predict movements of balls in simulated environments.

\paragraph{Graph Neural Networks (GNNs) for Learning Physics}

As deep structured models, GNNs allow learning useful representations of entities and relations among them, providing a reasoning tool for solving structured learning problems. Hence, it has found extensive use in physics prediction. Interaction network by Battaglia \textit{et al.}~\cite{battaglia2016interaction} and Neural Physics Engine by Chang \textit{et al.} \cite{chang2016compositional} are the earliest examples of general-purpose physics engines that depend on GNNs. These models do object-centric and relation-centric reasoning to predict the movements of objects in a scene. While they were successful in modeling dynamics of several systems such as n-body simulation and billiard balls, their models had certain shortcomings, especially when movements of objects have a chain effect on other objects (e.g., a pushed object pushes a group/sequence of objects it is contacting with) or when the objects are composed of complex shapes. These shortcomings can be partly handled by including a message passing structure within GNNs as done in the recent works such as~\cite{li2018propagation,mrowca2018flexible,li2018learning}. 
Most of these networks used simple neural networks for encoding object and relation information. Kipf \textit{et al.} \cite{kipf2018neural} showed that variational autoencoders could be used in encoding object and relation information, where their network was shown to encode object information directly from trajectories of the objects in an unsupervised way. 

Another approach was acquiring object information directly from images. Ye \textit{et al.} \cite{ye2019compositional} used image and detected the location of objects to predict the latent representation of the next time step. This latent representation was then decoded to create the image expected to be observed in the next time step. Watters \textit{et al.} \cite{watters2017visual} and van Steenkiste \textit{et al.} \cite{van2018relational} proposed hybrid network models which encode object information directly from images via CNNs and predict the next states of the objects via GNNs. Lately, these networks have been extended to handle even more complex environments. Sanchez-Gonzales \textit{et al.} \cite{sanchez2020learning} showed that GNNs could be used for learning particle-based simulations that consist of more than 1000 particles.

\paragraph{Effect Prediction in Robotics}

Action-effect prediction has been investigated using model-based approaches that use analytical models \cite{yu2016more, hogan2020feedback}, data-driven methods that use machine learning methods and hybrid methods that incorporate machine learning into analytical modeling \cite{zhou2018convex, kloss2017combining}. The effect prediction methods can be further divided into two categories depending on the number of involved objects. In order to deal with predicting action effects on single objects, object masks have been heavily used \cite{omrvcen2009autonomous,King2015, Haustein2015, kopicki2011learning}. Recently, Kopicki \textit{et al.} \cite{kopicki2017learning} proposed learning multiple motion predictor models for different shaped single objects, where a vision system selects a predictor depending on the context. Seker \textit{et al.}  \cite{SEKER2019173} investigated how changing object shapes affects low-level object motion trajectories and modeled it using CNNs and LSTMs.

In the context of end-to-end learning, Agrawal \textit{et al.} \cite{agrawal2016learning} trained forward and inverse models for learning how to poke an object to move it into a target position. This network uses latent vectors of CNN to train predictive models. The forward model tries to predict the latent representation of the final image using the current image,  and the inverse model took latent representations of both final and initial images to find the parameters of the poke action. Finn \textit{et al.}~\cite{finn2016unsupervised} proposed a convolutional recurrent neural network~\cite{xingjian2015convolutional} to predict the future image frames using only the current image frame and actions of the robot. Byravan \textit{et al.} \cite{byravan2017se3} presented an encoder-decoder like architecture to predict SE(3) motions of rigid bodies in depth data. However, the output images get blurry over time, or their predictions tend to drift away from the actual data due to the accumulated errors, making it not straightforward to use for long-term predictions in robotics. 

The previous data-driven methods that directly used object-centric representations cannot deal with multiple (any number of) objects and relations as the predictors have generally fixed input and output dimensions. End-to-end approaches can handle multiple objects as their inputs and outputs are images, however, the pixel-based prediction quickly accumulated, resulting in blurry long-term predictions. Recently, GNNs that can represent multiple objects in an object-centric way  have started being employed in robotics research as well. Janner \textit{et al.} \cite{janner2018reasoning} used GNNs to learn object representations from perception and physics prediction jointly. Ye \textit{et al.} \cite{ye2019object} learned object-centric forward models for planning and control. Their model takes object bounding boxes as input and learns future state prediction from object embeddings generated by CNNs. Tung \textit{et al.} \cite{tung20203d} similarly use object bounding boxes with GNNs for effect prediction and control. Paus \textit{et al.} \cite{paus2020predicting} used GNNs for action-effect prediction. Sanchez-Gonzales \textit{et al.} \cite{sanchez2018graph} have used graph networks as learnable physics engines in robotic setups. While previous GNN based robotic effect prediction models were successful in modeling physics, they largely overlook unknown or partial information. Our model can also handle more complex shaped objects by modeling them as a group of articulated simple shaped objects.

\paragraph{Parameter Estimation}

Wu \textit{et al.} \cite{wu2015galileo} proposed a deep approach for finding the parameters of a simulation engine that predicts the future positions of the objects that slide on various tilted surfaces. Zheng \textit{et al.} \cite{zheng2018unsupervised} used perception prediction networks, a type of graph neural network, for learning latent object properties from interaction experience to simulate system dynamics. 

In many scenarios, simply observing the scene may not yield enough information, and the robot may need to actively act on the environment to perceive more. In these cases, the robot can improve its perception by actions~\cite{bohg2017interactive}. Li \textit{et al.} \cite{li2018push} used recurrent neural networks to predict the center of mass from object mask and interaction experience. Xu \textit{et al.} \cite{xu2019densephysnet} used a deep learning architecture for learning object properties. In their settings, a robot slides an object from an inclined surface and cause it to collide with another. Using a sequence of dynamic interactions, they showed that their model could learn to predict object representations. Kumar \textit{et al.} \cite{kannabiran2019estimating} trained policy and predictor networks to estimate the mass distribution of articulated objects. They showed that their policy network improves the mass prediction capacity of the predictor network compared to the random policy. However, their approach was limited to articulated objects with a fixed number of parts.

In \cite{Sturm2009, Schmidt2014, Martin2016, Martin2019}, researchers also studied estimating the joint relations between objects for real-time tracking and prediction of the articulated motions in challenging interactive perceptual settings. These works, however, assume expert knowledge about the joint types and hard-code the corresponding transformation matrices~\cite{Schmidt2014}, candidate template models~\cite{Sturm2009}, specific measurement models~\cite{Martin2016, Martin2019} to detect kinematic structures. Our system assumes no prior knowledge about joint dynamics, and the robot learns the dynamics of categories purely from observations. Therefore, the learning dynamics of completely novel relation types is possible with our system. Exceptionally, in \cite{Sturm2009}, Sturm \textit{et al.} proposed to learn articulation dynamics from data; however, it was only realized on a single-pair of objects from a single articulation observation (garage door motion). Furthermore, these studies do not learn or predict how the pairs or chains of non-articulated touching objects would propagate the applied forces along the cluster/chain. In contrast, our system can predict the propagated effect on groups of touching non-articulated objects.

In our work, we verified the prediction and reasoning capability of the robot in use of tools that are composed of basic primitive shapes. While our main focus is not on decomposing objects into primitives, it should be noted that this topic has been studied in the literature. For example, Deng \textit{et al.} \cite{deng2020cvxnet} showed that from input images, objects can be decomposed into convex hulls. In addition, they showed that these convex hulls could be used for physics simulation. Similarly, Pashevich \textit{et al.} \cite{pashevich2020learning} proposed a framework that can propose different part sets where objects can be divided into, and then reconstruct the divided object in the real world with a robot using the available primitives in the workspace. 

%% file: sections/method.tex
\section{Proposed Framework}

\begin{figure*}[t]
    \centering
    \includegraphics[width=\linewidth]{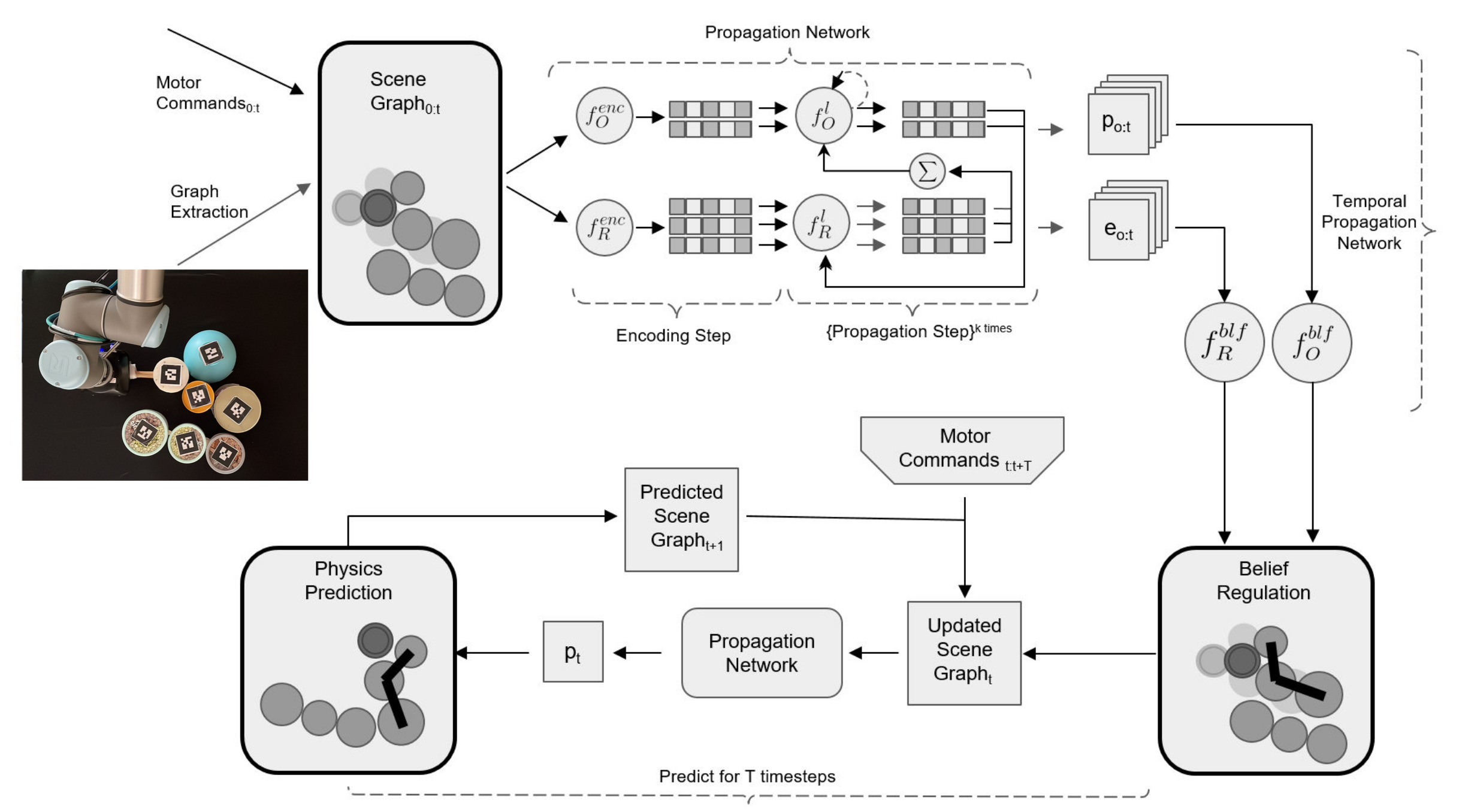}
    \caption{Our framework extracts object- and relation-centric latent representations from the current physical scene. The latent representations are initially used to update unknown parameters of the scene graph, then with the planned motor commands, they are used for predicting future motion of the manipulated objects. \label{fig:complete_system}
    } 
    
\end{figure*}

we propose methods and framework that are capable of learning object- and relation-centric representations for different physical scenes. These representations can be used in a set of various tasks. In this work, we designed our framework around solving two complementary tasks, namely \emph{belief regulation} and \emph{physics prediction}. Figure~\ref{fig:complete_system} shows a graphical illustration of our framework. First, object- and relation-centric representation for each object and their object-object relations are learned using propagation network. By giving these representations to RNN networks, our framework finds unknown object and relation parameters and acquires an updated graph of the scene. By passing the updated scene graph and future robot actions to the same propagation network, our framework predicts the future motion of the manipulated objects by chaining the effect predictions. In the rest of the section, more technical details will be provided.

\subsection{Preliminaries} 

\paragraph*{Physical System as a Graph}

From a physical system with multiple interacting objects, we form a graph $G=\left\langle O, R\right\rangle$  where each object $O$ is represented by the nodes (of cardinality $N^o$) $O = \{o_i\}_{i=1:N^o}$  and the relations $R$ between objects such as a contact or a joint are represented by the edges (of cardinality $N^r$) $R = \{r_{k}\}_{k=1:N^r}$ of the graph. 

\paragraph*{Representing Push Manipulation Tasks}

We are interested in representing the push manipulation task as a robot interacting with an object clutter. The clutter could contain many objects that may have different parts with different mass distributions, objects with possible articulations, etc. We plan to represent such a system with the aforementioned graphs $G=\left\langle O, R\right\rangle$. 

Each node $o_i= \left\langle {x_i},{a^o_i}\right\rangle$ store object or part vectors, where ${x_i} = \left\langle {q_i},{\dot q_i}\right\rangle$ is the state of the object $i$, with its pose ${q_i}$ and velocity ${\dot{q_i}}$. $a^o_i$ stands for object properties such as shape or mass. Between each $i$, $j$ node pair, there is an edge $r_k= \left\langle {d_k},{s_k},{a^r_k}\right\rangle$ that represents object-object relations where ${d_k}=q_i-q_j$ stands for displacement vector, ${s_k}=\dot q_i-\dot q_j$ stands for velocity difference, and ${a^r_k}$ corresponds to properties of relation $k$ between objects $i$ and $j$. 

\begin{figure*}[t]
    \centering
    \includegraphics[width=0.8\linewidth]{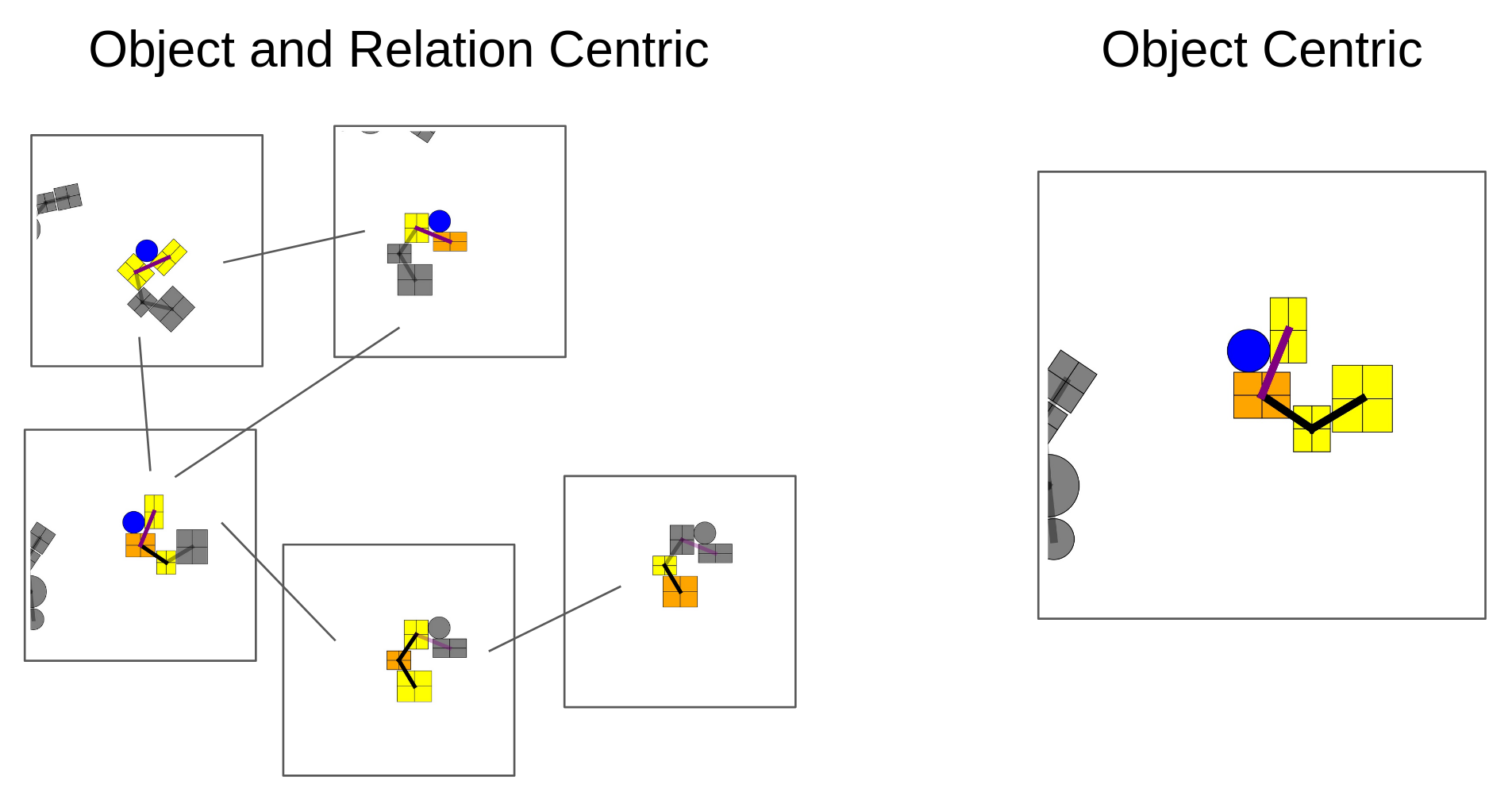}
    \caption{Comparison between object-centric vs object- and relation-centric representations. The representation on the left allows network to capture object details in a more compositional way, allowing network to propagate action-effects between objects and predicting action-effects of each object more accurately. 
        \label{fig:centric_comparison}
    }
\end{figure*}

Previously, push-effect prediction tasks were often represented with object-centric image-based representations. Compared to these representations, our graph based representation can represent scene in both object- and relation-centric way. The visualization of this is shown in Figure~\ref{fig:centric_comparison}. This inductive bias allows our network to capture scene dynamics more accurately and efficiently. For object-centric representations, action-effect for each object has to be calculated separately which may more often result in inconsistent predictions.  

\paragraph*{Representation of Robot} We propose representing the end-effector of the robot as a part of the graph. For this, a robot flag and a control vector that shows how the end-effector will move in the next step are used.

\paragraph*{Leveraging Graph Representation}

For this work, our representation covers cylinders, cuboids, and objects that can be represented with the combination of two. Objects in the scene are represented with their shape, state, and other object features such as mass. Shape of objects are represented with their dimensions (the radius for cylinder and edge lengths for cuboid) and their orientations. Orientations of objects are represented with vector $[cos(\theta),sin(\theta)]$ for 2D cases, and with quaternions for 3D cases. Unlike previous work\cite{Sturm2009, Schmidt2014, Martin2016, Martin2019}, the system has no prior information about how joints behave, and the articulation dynamics are left for the network to learn. 

\subsection{Physics Prediction}

\paragraph*{Propagation Network} 
We used propagation network as a base for learning object- and relation-centric representations. In this network, first, the state of each object and the relations between them are encoded separately. This step is shown in Figure~\ref{fig:complete_system} (Encoding-Step).
The encoding process is achieved by use of $f^{enc}_R$ and $f^{enc}_O$ encoders where former process relation features $r_{k,t}$, while the latter process the object features $o_{i,t}$. $c^r_{k,t}$ and $c^o_{i,t}$ are the latent encodings of the objects and the relations.

\begin{eqnarray}
    c^r_{k,t} = f^{enc}_R\left(r_{k,t}\right), \quad k=1\ldots N^r\\
    c^o_{i,t} = f^{enc}_O\left(o_{i,t}\right), \quad i=1\ldots N^o
\end{eqnarray}

\begin{figure*}[t]
    \centering
    \includegraphics[width=0.95\linewidth]{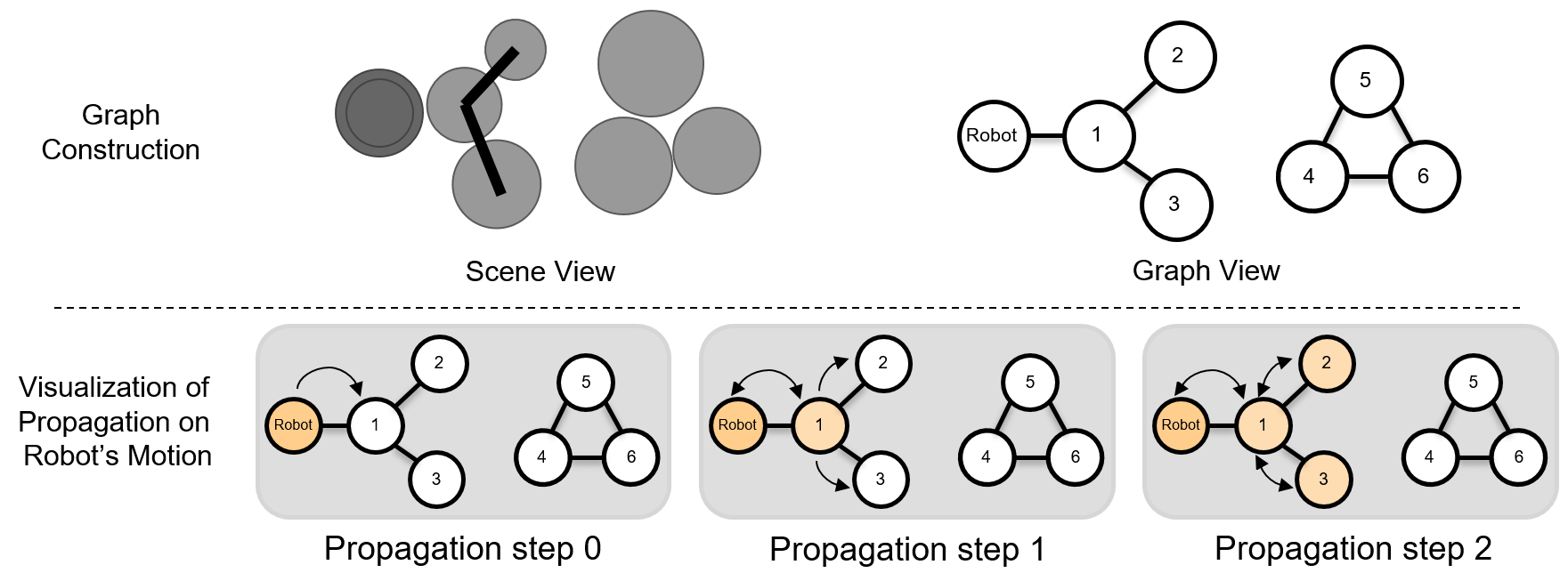}
    \caption{
        This illustration shows how the graph of the scene is constructed and how the force emerging from robot end-effector motion is passed to the faraway objects. After graph construction, each node holds state information of their corresponding objects, including the robot. Considering how state information of robot is passed, in the first propagation step, it is passed to nodes of objects that contact the robot end-effector. In the second propagation step, via nodes of objects that the robot initially contact, this state information is passed to nodes of non-contacted objects.
        \label{fig:propagation_illust}
    }
\end{figure*}

Next, the network incorporates interactions between objects and propagations of these interactions between non-neighbor objects (e.g., force transmission between non-contacting objects) into object and relation latent vectors. This step in shown in Figure~\ref{fig:complete_system} (Propagation Step). For this, $c^r_{k,t}$ and $c^o_{i,t}$
are passed to propagator functions $f^{l}_R$ and $f^{l}_O$ respectively for estimating propagation latent vectors $e^{l}_{k,t}$ for relation $k$ and $p^l_{i,t}$ for object $i$, for each propagation step $l$ at time $t$. Using these functions in subsequent propagation steps allow for nodes and edges to accumulate propagated information from nodes and edges connected to them in $e^{l}_{k,t}$ and $p^l_{i,t}$.  
\begin{eqnarray}
    e^{l}_{k,t} = f^{l}_R\left(c^r_{k,t},p^{l-1}_{i,t},p^{l-1}_{j,t}\right), \quad k=1\ldots N^r\\
    p^l_{i,t} = f^{l}_O\left (c^o_{i,t},p^{l-1}_{i,t},\sum_{k \in \mathcal{N}_i} e^{l-1}_{k,t}\right ), \quad i=1\ldots N^o
\end{eqnarray}
where $\mathcal{N}_i$ stands for set of relations object $i$ is part of.

Effect propagation allows network to pass information between non-connected objects, and it benefits our framework in two important ways. Firstly, it allows force transmission when the robot pushes objects towards another one, effectively pushing both objects while contacting only one of them. Secondly, it allows mass and friction feedback between objects (e.g., when a light object is pushed towards a heavy object, the light object will not move the heavy objects in the push direction, but instead its motion will be shifted toward light or left side.). Figure~\ref{fig:propagation_illust} shows a simple illustration of how the robot initiates a chain of interaction and how force applied by the robot end-effector propagates. In initial propagation step, force that emerge from motion of robot is passed to contacted objects and in second propagation step, this force propagates to non-directly interacted objects. How many subsequent propagation steps to apply can be chosen based on the difficulty of the task.

Resulting $e^{l}_{k,t}$ and $p^l_{i,t}$ well represent the objects and their relations in the graph and can be further passed to other networks for physics prediction and belief regulation.

\paragraph*{Physics Prediction} For each object, the latent vector $p^l_{i,t}$ can be used to predict the next  state of object $x_{i,t+1}$. 
Given states of the objects in time $t$, our framework can be used for predicting the trajectory rollout of objects between time $t$ and $t+T$ by chaining its estimates, using the predictions as an input for estimating subsequent states of objects. 

\subsection{Belief Regulation}

\paragraph*{Temporal propagation network} 

We propose a temporal propagation network to estimate and correct object and relation properties over time. The propagation network is augmented with long short-term memory (LSTM) networks to regulate object and relation beliefs.
Network illustration is shown in Figure~\ref{fig:complete_system} (Temporal Propagation Network).
In temporal propagation network,  sequence of propagation latent vectors $e^{l}_{k,t}$  and $p^l_{i,t}$ are passed to LSTM-based encoder functions $f^{blf}_R$ and $f^{blf}_O$. In this way, the temporal propagation network estimates and corrects object and relation properties by considering their overall state history during the robot execution.

\begin{eqnarray}
    o'_{i,t} = f^{blf}_R\left(p^{L}_{i,t},o'_{i,t-1}\right), \quad i=1\ldots N^o\\
    r'_{k,t} = f^{blf}_O\left(e^{L}_{k,t},r'_{k,t-1}\right), \quad k=1\ldots N^r
\end{eqnarray}

\paragraph*{Belief Regulation} Belief Regulation module can continuously regulate beliefs regarding objects and relations states ($o_{i,t}$ and $r_{k,t}$). These beliefs can then be used in physics prediction to compensate for errors that arise from unknown or partial information regarding the scene. This will allow our network to close the gap between its physics predictions and reality.

\paragraph*{Weight Sharing}
After training the propagation network for physics prediction, learned weights can be reused in belief regulation, preventing the framework from having to learn two separate networks. This decreases the number of parameters by about thirty percent. As we show in our experimental analysis, the representation used with physics prediction well represents the environment and can be used in transfer learning without affecting the system performance.

%% file: sections/experiments.tex
\section{Experimental Setups}
\label{sec:Experimental Setups}

In this section, we explain the details of the experimental setups that are designed to evaluate how our model can be used for predicting object properties, relations between objects, and future object trajectories. 

\subsection{Robotic Setup}

Experiments are conducted with a 6 DoF UR10 robot arm with a cylinder shaped object attached to its end-effector both in simulation and real-world. For simulation experiments, CoppeliaSim~\cite{rohmer2013v} with Pyrep toolkit~\cite{james2019pyrep} is used. For demonstrating prediction capacity of our framework, two different object setups, namely \textit{Multiple Parts Setup} and \textit{Different Masses Setup}, are defined. The former setup includes a diverse set of interactions in the form of joints and is designed with the aim of showing the full capacity of our framework. As the physical effects of object parameters are limited in the former setup, the latter setup is designed with the aim of showing the performance of our framework in setups where effect variation result from object parameters. In these setups, edges between objects are dynamically created as objects approach each other. As the robot interacts with the objects in the environment, only a certain subset of objects will be in the same sub-graph of the robot (This can be seen in Figure \ref{fig:propagation_illust} graph view.), and accordingly, this allows the system to encounter sub-graphs with a different number of objects and relations. 

\begin{table}[t!] 
    \caption{Explanations of the joint types and their effects.}
    \begin{tabular}
        { llc } \hline Example setup   & Effect of action & Outcome Explanation \\
        \hline 
        \hspace{0.2mm}
    \parbox[c]{0.14\linewidth}{
        \fbox{\includegraphics[width=\linewidth]{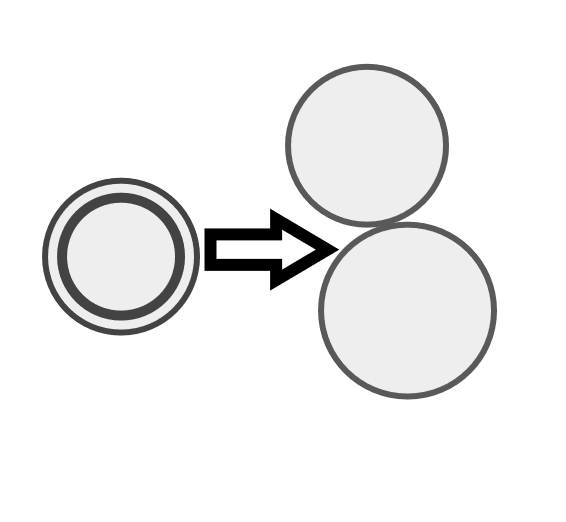}}}     & 
        \parbox[c]{0.1806\linewidth}{
        \fbox{\includegraphics[width=\linewidth]{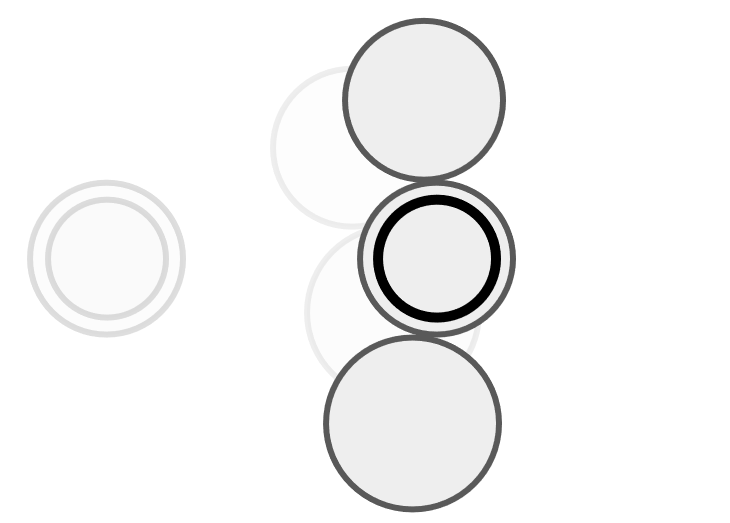}}}      
        \vspace{1mm} 
        & \parbox[c]{0.47\linewidth}{\textit{No joint:} The objects would move independent of each other as they are separated by the gripper.}   \\
        \hspace{0.2mm}
        \parbox[c]{0.14\linewidth}{
        \fbox{\includegraphics[width=\linewidth]{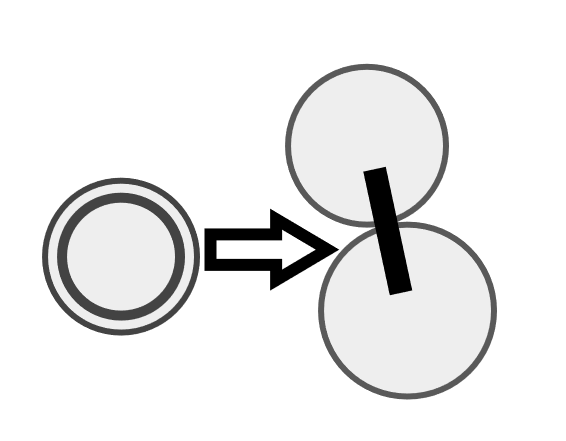}}}     & 
        \parbox[c]{0.1806\linewidth}{
        \fbox{\includegraphics[width=\linewidth]{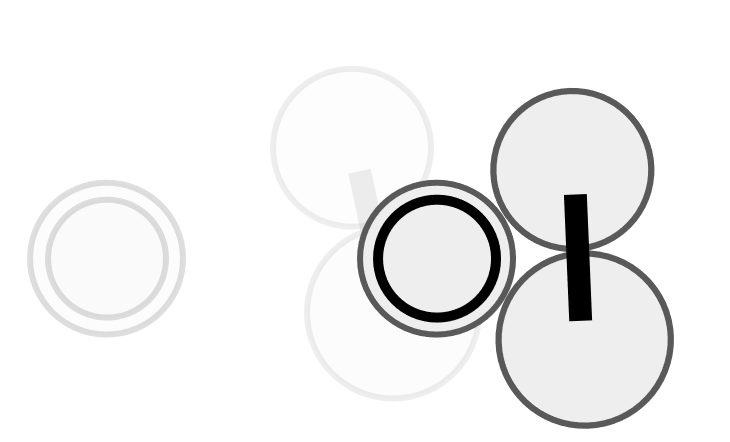}}}      
        \vspace{1mm} 
        & \parbox[c]{0.47\linewidth}{\textit{Fixed joint:} The objects would move together with the end-effector of the robot.
        }    \\
        \hspace{0.2mm}
        \parbox[c]{0.14\linewidth}{
        \fbox{\includegraphics[width=\linewidth]{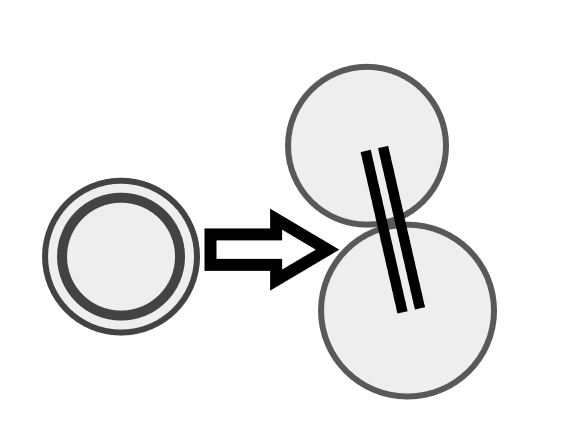}}}     & 
        \parbox[c]{0.1806\linewidth}{
        \fbox{\includegraphics[width=\linewidth]{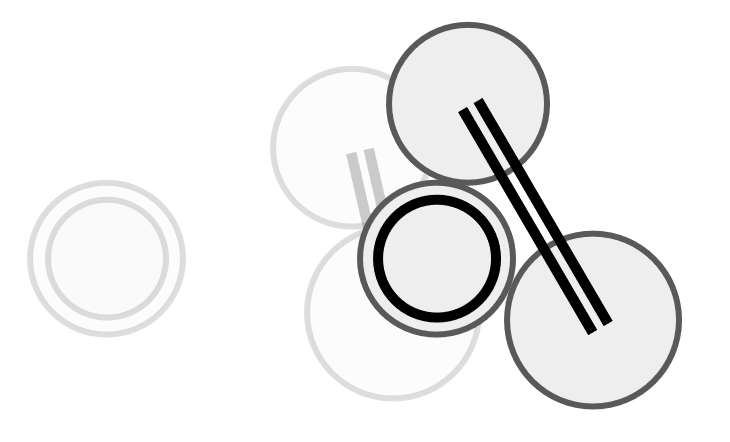}}}      
        \vspace{1mm} 
        & \parbox[c]{0.47\linewidth}{\textit{Prismatic joint:} The object below would move in linear line along the direction between the above object to below object. 
        }   \\
        \hspace{0.2mm}
        \parbox[c]{0.14\linewidth}{
        \fbox{\includegraphics[width=\linewidth]{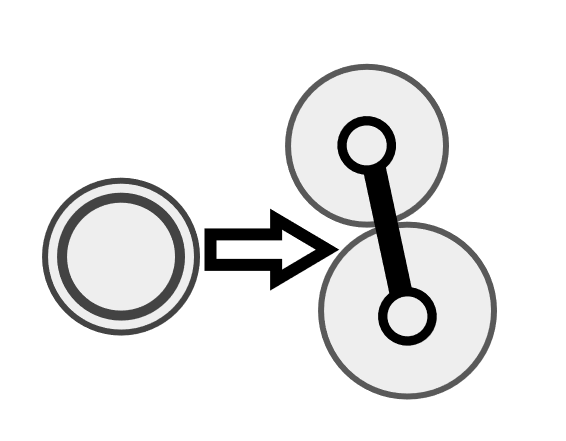}}}     & 
        \parbox[c]{0.1806\linewidth}{
        \fbox{\includegraphics[width=\linewidth]{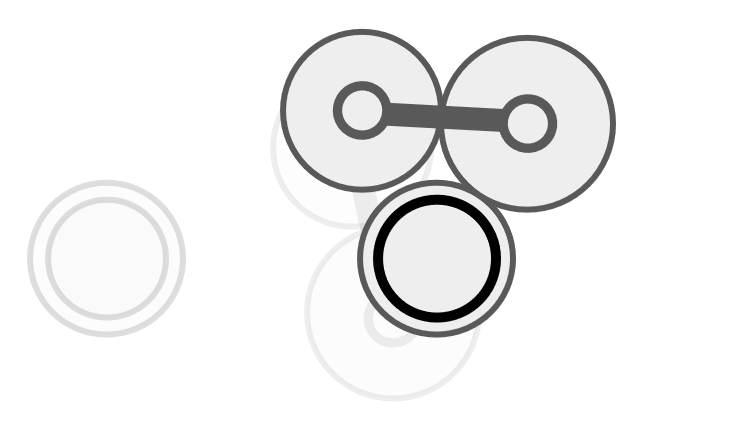}}}      
        \vspace{1mm} 
        & \parbox[c]{0.47\linewidth}{\textit{Revolute joint:} Both of the objects would move, but as the end-effector mainly contacts object below, the robot will rotate the object below around the object above.
        }   \\
        \hline
    \end{tabular}
    \vspace{1ex}

     {\raggedright Note: Objects and the robot are shown with single-edged and double-edged circles respectively, and the lines between objects represent different joint types. The arrow shows how the robot end-effector will move.\par} 
    \label{table:joints}
  \end{table}

\textit{Multiple Parts Setup:} This setup consists of a group of articulated objects where our framework should learn dynamics of objects, including cylinders and cuboids, with complex spatial relations between them. The objects may be connected to each other through three different joint relation types, namely \emph{fixed}, \emph{revolute} and \emph{prismatic} joints, or they may have no joint connections between them (\emph{no-joint}). The Illustration of these joint relations and their explanations are shown in Table~\ref{table:joints}.

\textit{Different Masses Setup:} This setup consists of differently massed cylindrical objects where masses of objects have an effect on their future motion. From the motion trajectories of the objects, our framework should be able to predict their masses. The masses are sampled from three intervals: $ 0.2 - 0.5\ kg$, $ 1.0 - 2.0\ kg$, $ 8.0 - 10.0\ kg$, representing light, normal and heavy objects, respectively.

For both of these setups, we generated datasets containing 30,000 training and 1000 validation trajectories with 9 objects. Since it is hard to exactly tune end-effector velocity to match real-world, end-effector velocity of the robot is changed between different trajectories so that it can generalize to different values. For testing the generalization capacity of the network to changing number of objects, we used trajectories consisting of 9, 6, and 12 objects, each with 1000 trajectories. 

\subsection{Implementation Details}

\paragraph*{Generation of Graph} For each object in the scene and close-by object pair, a node and two directed edges (a receiver and a sender) are created. To make the system position and orientation invariant, object position and orientations are not included in the node features. Instead, for each object-object relation, the pose of the object on the sender side of the relation is encoded with respect to frame of the object on the receiver side of the relation. After the motion of an object on its own frame is predicted, it is transformed back to the global frame.

\paragraph*{Network information} 
 $f^{enc}_O$ is a two 256-dim hiddenlayer MLP, and $f^{enc}_R$ is a three 256-dim hidden layer MLP.  $f^{l}_O$ and $f^{l}_R$ are MLPs with 256-dim single hidden layer. $f^{l}_O$ and $f^{l}_R$ are chosen to have a low number of layers since these network called multiple times successively and therefore more costly to use than $f^{enc}_O$ and  $f^{enc}_R$. Finally, $f^{blf}_O$ and $f^{blf}_R$ are LSTM with 256 neurons.  For physics prediction, outputs of $f^{l}_O$ is given to an MLP with one hidden layer and one linear layer to predict velocity (${\dot{q_i}}$) of each object; and for belief regulation, outputs of $f^{blf}_O$ and $f^{blf}_R$ are given to an MLP with single linear layer to predict object masses and joint relations.
 
In the belief regulation module, as more interaction experience is acquired, the framework is expected to have higher accuracy in identifying initially unknown parameters of the environment. For this reason, the loss function is scaled in a way that further time-steps have a higher loss value compared to earlier time-steps. Besides, to make networks predictions smooth and preventing them from oscillating between different outcomes, outputs of $f^{blf}_O$ and $f^{blf}_R$ are regularized by applying MSE loss between latent vectors of successive time-steps.

The network is trained with 16 batch-size and 3e-4 learning rate using Adam optimizer\cite{kingma2014adam} with AMSgrad\cite{reddi2019convergence}.
The learning rate is reduced by 0.8 when the validation error stopped decreasing for a window of 20 epochs. Networks are trained for 1000 epochs. The physics prediction module is trained with epochs of 10,000 batches of randomly sampled time-steps, and for the training belief regulation, 200 batches of randomly sampled trajectories from the training scenes are used.

First, our network is trained on physics prediction. After the training is complete, the weights of the shared part of the network are frozen, and then the belief regulation module is trained. To increase the performance of physics prediction, we used scheduled sampling\cite{bengio2015scheduled}. 
Using Nvidia P100 GPU, the physics prediction and belief regulation modules are trained for two and one days respectively.  

\section{Results}
\label{sec:Results}

For quantitative analysis, our framework is evaluated in joint prediction and mass prediction tasks.  For the relation prediction case, our results are compared with PropNets with three different relation assignment strategies.
\begin{enumerate}
	\item \textbf{Oracle} This relation assignment strategy utilizes ground-truth relations. In the ideal case, as more interactions are observed, the performance of our framework should approach to oracle.
	\item \textbf{No-Joint} This relation assignment strategy assumes there are no joints in the scene.
	\item \textbf{All-Fixed} This relation assignment strategy assumes a fixed joint between every contacting object pairs.
\end{enumerate}

\subsection{Quantitative Analysis in Multiple Parts Setup}

\begin{figure}[!t]
    \centering
    \includegraphics[width=0.9\linewidth]{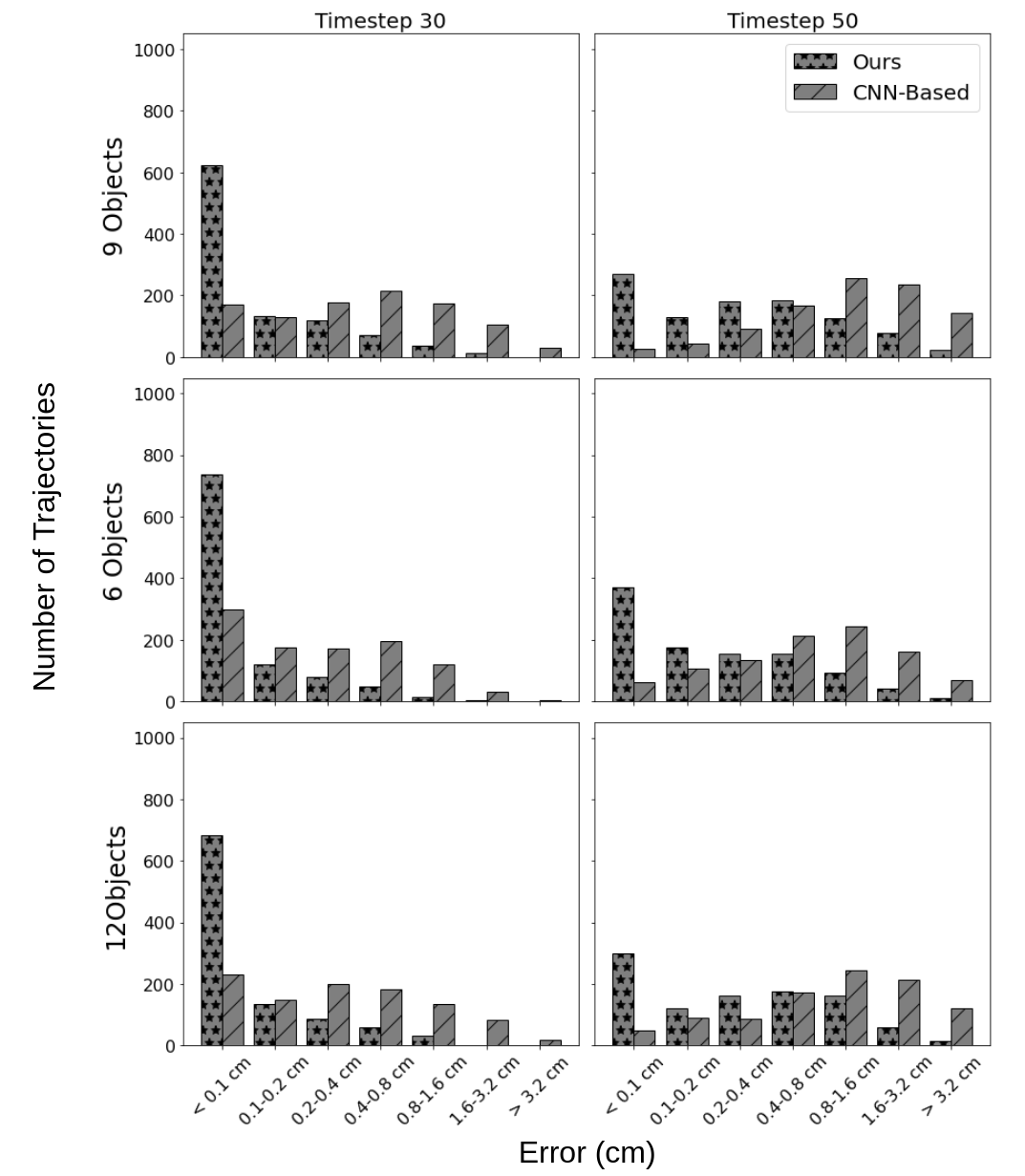}
    \caption{Physics prediction results on articulated object environments. Error distribution of our network is skewed toward lower error, while CNN-based architecture that uses object-centric images has error distribution skewed towards higher error.}
    \label{fig:physics_articulated}
\end{figure}

For evaluating the physics prediction module, we compared our graph-based architecture with an CNN-based physics prediction architecture that takes rendered object-centric images of pushed objects and predicts the displacement vector. 
Both architectures are tested with the oracle relation assignment strategy in \textit{multiple parts setup}.  In this setup, while collecting each trajectory, the robot executes 9 linear pushes of 30 cm, contacting with a most diverse set of objects. For both architectures, outputs of physics predictions are chained to predict multiple time-step trajectory roll-outs (i.e., essentially simulating the environment with network predictions). These trajectory roll-outs are used in evaluation. Figure~\ref{fig:physics_articulated} presents the performance in scenarios with different number of objects. Comparing error values, it can be seen that both architectures are able to generalize to different number of objects. However, our architecture performs significantly better than then CNN-based one as errors for our network are skewed towards low values while CNN-based architecture one has errors skewed towards high values. Regardless, as the length of predicted trajectory roll-outs increase, the errors in higher number of trajectories accumulate. This results in trajectories to drift away from the ground truth. For our architecture, in Figure~\ref{fig:physics_articulated} on the left, as the roll-out length is shorter in each environment setup, more than 600 trajectories have lower mean error than $0.1\ cm$, and most of the remaining trajectories have a lower mean error than $0.4\ cm$. On the right, the roll-out length is longer, and less than 400 trajectories have a lower mean error than $0.1\ cm$. Besides for the rest of the trajectories, there are more trajectories in high mean error bins.

\begin{figure}[!t]
    \centering
    \includegraphics[width=0.96\linewidth]{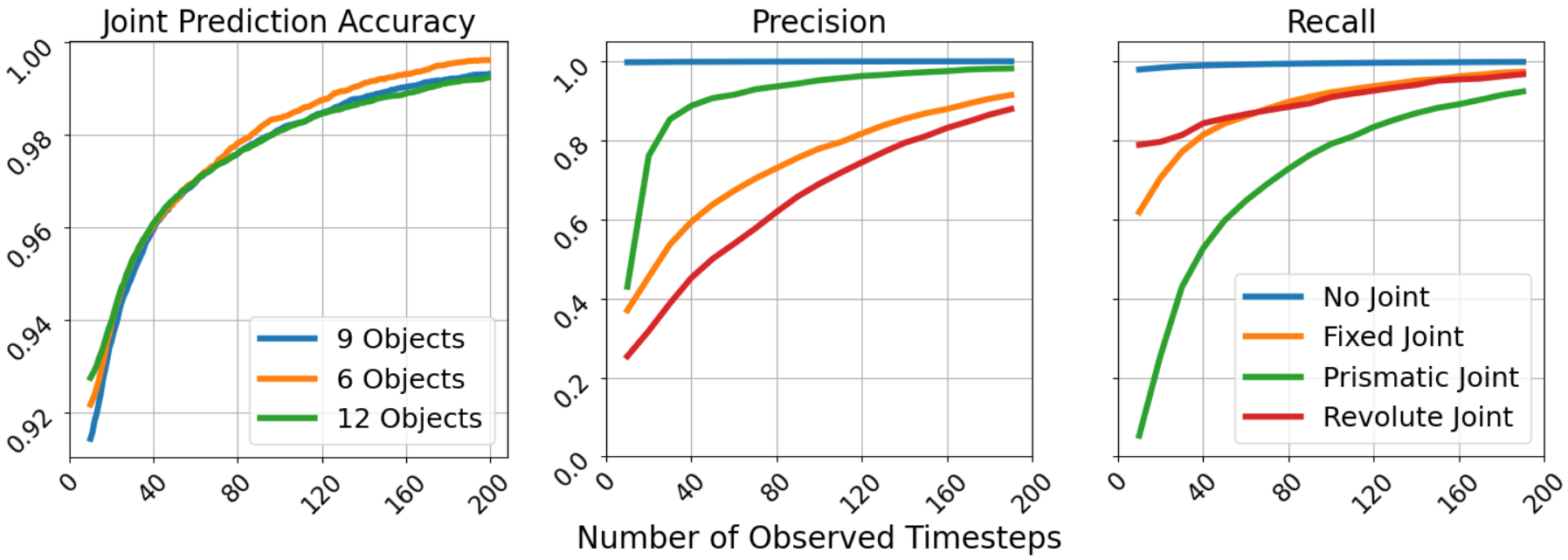}
    \caption{Belief regulation results on articulated object environments.}
    \label{fig:belief_articulated}
\end{figure}

Next, the belief regulation module is evaluated on prediction of joint relations.  As shown in Figure~\ref{fig:belief_articulated}, as the robot interacts with the objects and higher number of observation data is acquired, our network becomes better at predicting the joint relation types more accurately. The joint prediction plot in Figure~\ref{fig:belief_articulated} shows that our method performs similarly independent of the number of objects used due to the underlying graph structure. In the same figure, on precision plot, no joint (blue) and prismatic joint(green) lines show that networks are good at identifying whether there is a joint between two objects and whether this joint is prismatic.  Compared to the prismatic joint, the model is more likely to make erroneous predictions on whether a joint is fixed or revolute. This is likely because without interaction experience, it is easier for network to mix these two joints. Nonetheless, from the recall plot, we can see that the model can correct its predictions on fixed and revolute as it observes more robot interactions. From both precision and recall plots, the model abstains from predicting a joint prismatic unless it is certain. This may be because prismatic joint dynamics are similar to no-joint dynamics unless robot gains enough observations about objects the joint are connected to.

\begin{figure}[!t]
    \centering
    \includegraphics[width=0.9\linewidth]{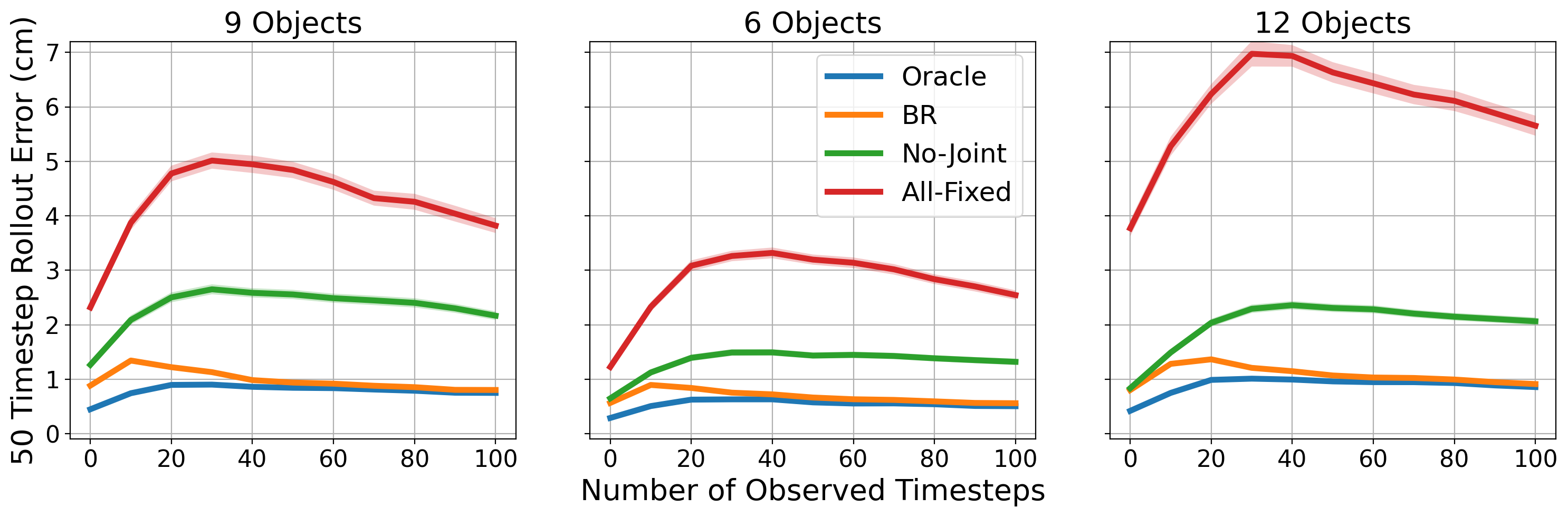}
    \caption{ Results of coupled system on articulated object environments.}
    \label{fig:full_system_articulated}
\end{figure}

Finally, the coupled results of the physics prediction and the belief regulation modules can be seen in Figure~\ref{fig:full_system_articulated}. The lines show the mean errors, and the shaded regions show the standard error. As expected, physics prediction done with no-joint and all-fixed relation assignment strategies performed poorly. This is because these relation assignment strategies do not learn from interactions. As the number of observed time-steps increases, the mean error of the coupled modules decreases and eventually in 40 time-steps, it reaches to the mean error of the physics prediction of the oracle system that has access to ground-truth joint relations.

\subsection{Quantitative Analysis of Belief Regulation for Mass Prediction}
\begin{figure}[!t]
    \centering
    \includegraphics[width=0.8\linewidth]{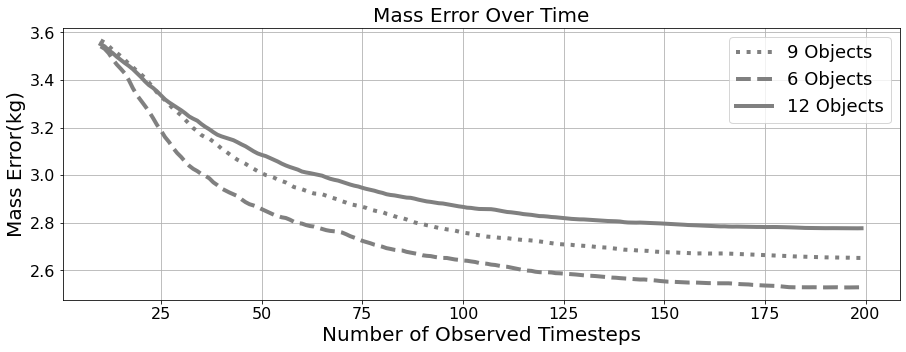}
    \caption{Belief Regulation results on mass prediction. As more motion is observed in the scene, mass prediction error decreases, but eventually converges to about ~$2.7\ kg$ mean error.
    }
    \label{fig:belief_mass}
\end{figure}

We design \textit{different masses} experimental setup for further testing the object-centric prediction capacity of our framework. In this setup, in each trajectory, the robot executes a total of 3 linear pushes of 30 cm, scattering objects as much as possible. In this experiment, our framework should predict object masses, and as the robot acquires more observations, it should improve its mass prediction accuracy further.
Mean errors for mass prediction is shown in Figure~\ref{fig:belief_mass}. Considering the distribution masses, our model manages to decrease mass errors over time as it acquires more observations. However, the predictions seem to not go below a certain value. This may be because the robot has limited interaction with the objects in the scene, and this limits the capacity of the model to predict masses of objects correctly.

\begin{figure}[!t]
    \centering
    \includegraphics[width=0.9\linewidth]{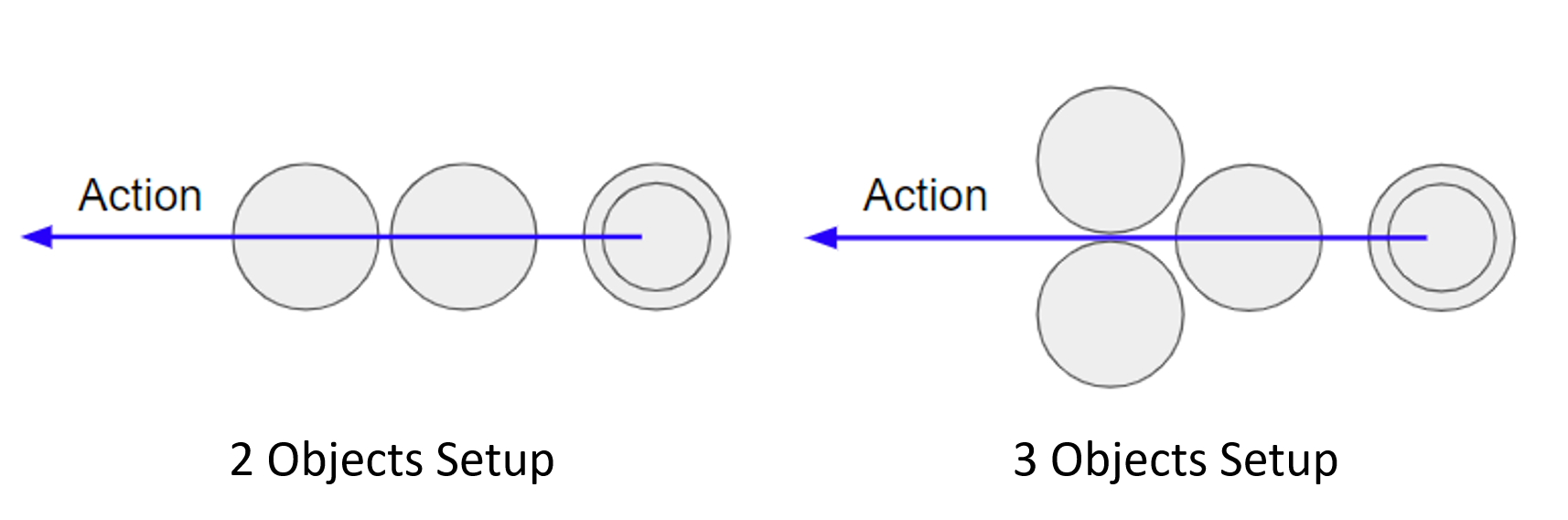}
    \caption{Visualization of controlled environment setups for mass prediction. In these configurations, object masses are changed between different runs while keeping robot motion and object shapes the same. }
    \label{fig:controlled_envs}
\end{figure}

\begin{figure}[t]
    \centering
    \includegraphics[width=0.75\linewidth]{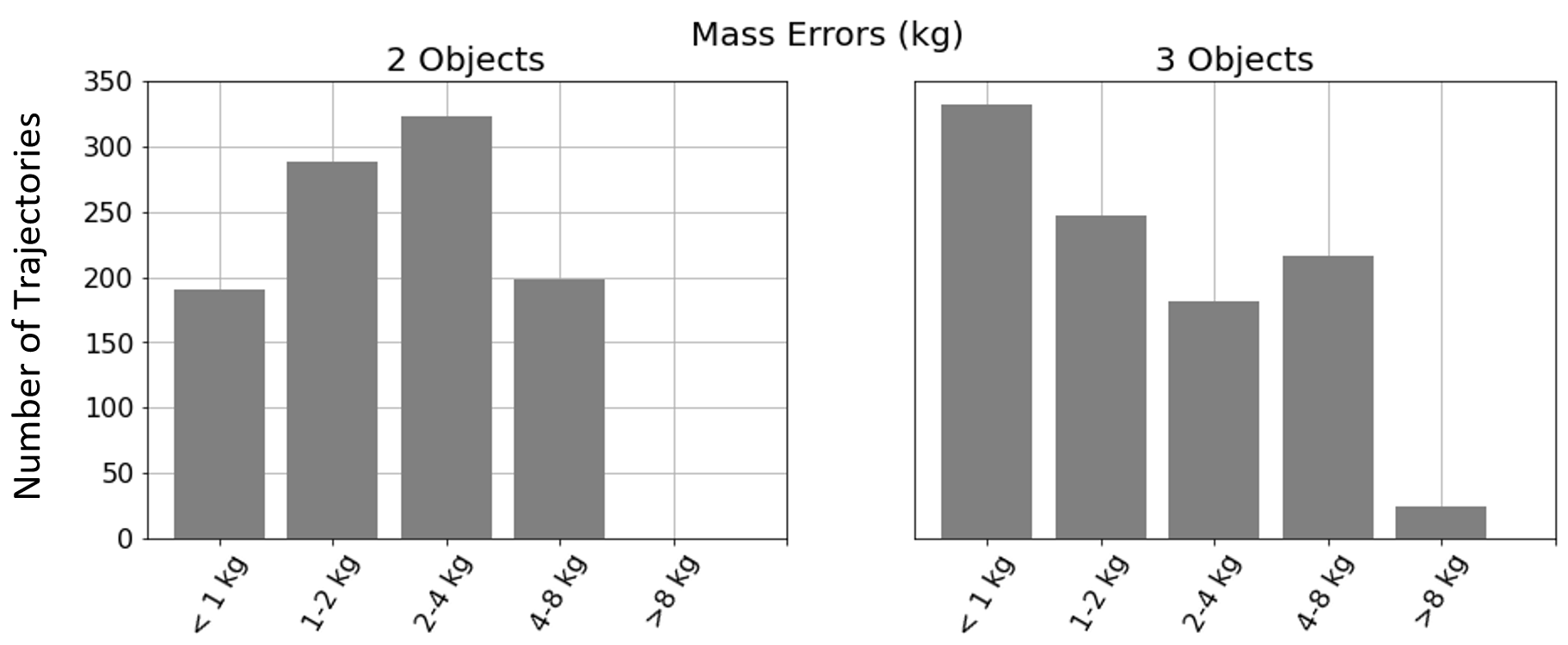}
    \caption{Mass prediction results in controlled environments. In many cases, our model acquires low error, however there are still many cases that have high error.}
    \label{fig:mass_error_2_3}
\end{figure}

\begin{figure}[t]
    \centering
    \includegraphics[width=\linewidth]{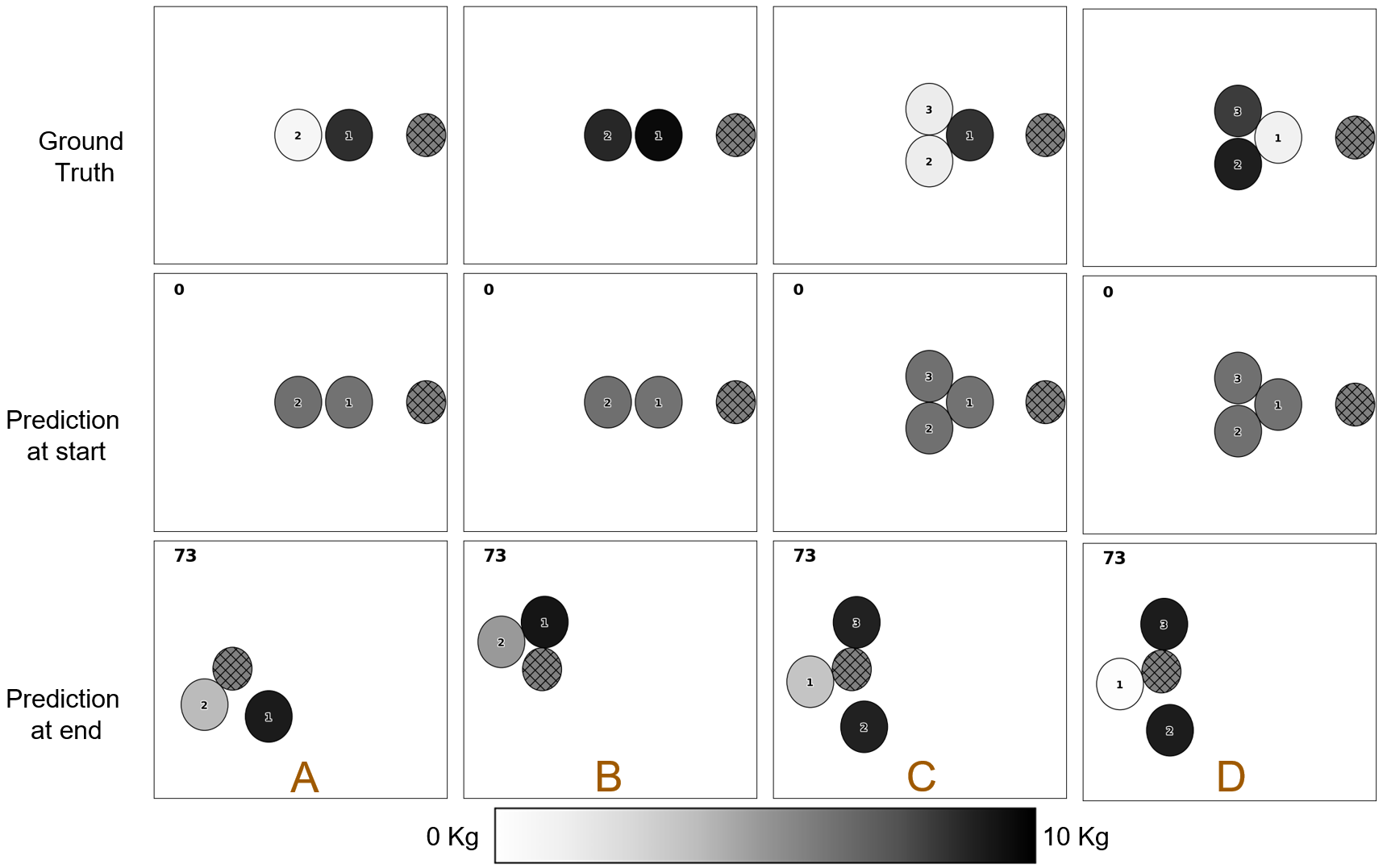}
    \caption{Mass predictions for two very close observations. The same observations are acquired from scenes with two different mass configurations, and our framework could not differentiate between the two. Our framework makes the same mass prediction for both; one of them is correctly predicted, while the other is not. }
    \label{fig:ambigous}
\end{figure}

To further analyze the performance of our system in mass prediction, we prepared two controlled environment test setups to examine why mass error does not decrease below a certain value.  These setups can be seen in Figure~\ref{fig:controlled_envs}. In these setups, we only change the mass of objects while setting robot action, initial positions  of objects, and shapes of objects same. The robot manipulates each object, so it should be possible for the network to predict mass if it is predictable. The results obtained in the these controlled settings are provided in  Figure~\ref{fig:mass_error_2_3}. Considering the mass distribution of the objects, the first two bars of both plots show that our framework predicts light and medium within their cluster correctly half of the time. The third and fourth bin shows that our framework sometimes mixes light and medium objects and medium and heavy objects. For three objects, the fifth bin shows that our framework mixes light and heavy objects in rare cases\footnote{
Videos of the results are available at project page. 
} 
A number of representative correct and incorrect predictions are provided in Figure~\ref{fig:ambigous}. We investigated setups where the network made high-error in mass predictions and observe that there are cases where different objects mass configurations having same object motions. Figure~\ref{fig:ambigous}C and Figure~\ref{fig:ambigous}D, the robot observes very similar trajectories with $0.15$ cm difference between them, despite the interacted objects having very different masses. In these scenes, the network makes very similar predictions. However, only in the former scene, it is correct.

\subsection{Qualitative Analysis - Tool Usage}

We design a tool manipulation and planning experiment. Given a goal position, the aim is to select the best tool and action sequence to bring a given object to the goal position using the corresponding tool. In addition, this experiment aims to show generalization capacity of our framework by transferring representation and the network trained in \textit{multiple parts setup} for modelling novel tools that are not encountered in the training distribution. 

\begin{figure}[t]
    \centering
    \includegraphics[width=0.6\linewidth]{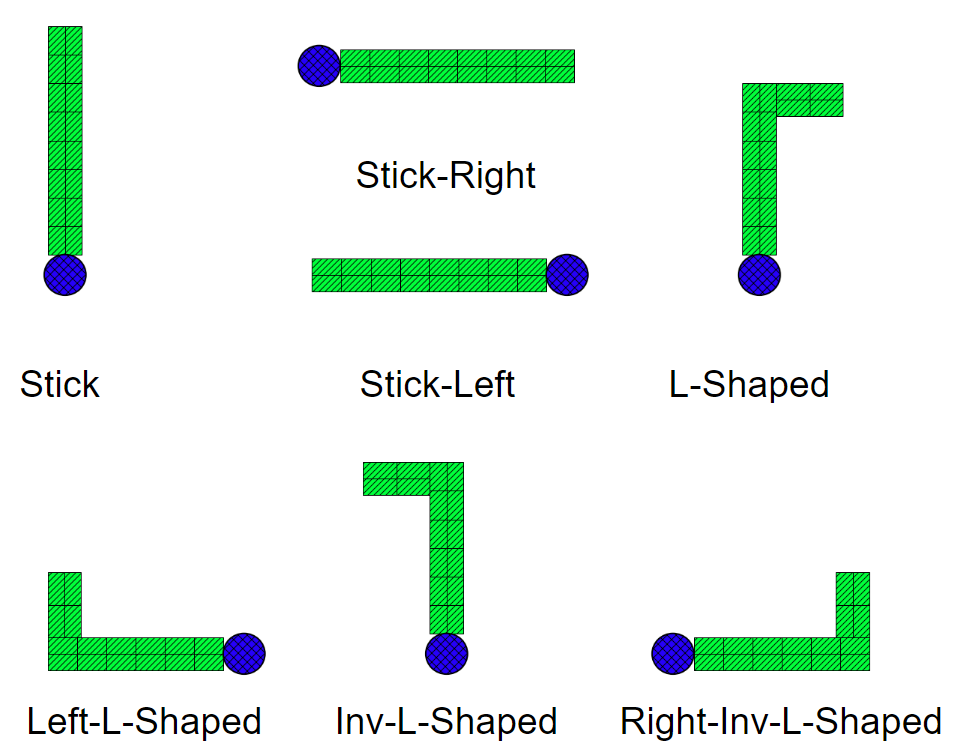}
    \caption{Tools used In tool selection and planning experiments.}
    \label{fig:used_tools}
\end{figure}

In this experiment,  stick, L-shaped tool, inv-L-shaped tool, and their various configurations are used as shown in Figure~\ref{fig:used_tools}. These tools are represented as multi-part objects composed of cuboids and fixed joints, and are attached to robot end-effector. The robot uses linear pushes in principal directions to manipulate the object on the table. In these actions, tool motion is modeled kinematically and not updated from the network prediction. Please note that a new network is not trained and the  results  obtained  by  the previously  trained  network are reported.

In each test case, the robot should select one of the available tools and apply three pushes of 20 cm in principle directions to move an object to a given goal position. To make all test cases feasible, goal points are generated through simulation. More specifically, 24 uniform initial positions are generated from $-0.7 \leq x \leq -0.1$ and $-0.5 \leq y \leq 0.5$ for. Then, on each initial position, a cylindrical object is generated, and all possible action sequences are applied using each of the tools. The final positions of objects are recorded. These final positions are filtered where if a final position of object is less than  5 cm away from its initial position, it is removed. Besides,  if the difference between any two initial and final position pair is lower than 5 cm, one of them is removed as well.  In this way, a dataset for tool and action selection that contains 166 completely diverse solvable initial and final position pairs are generated. 

\begin{figure}[!t]
    \centering
    \includegraphics[width=0.45\linewidth]{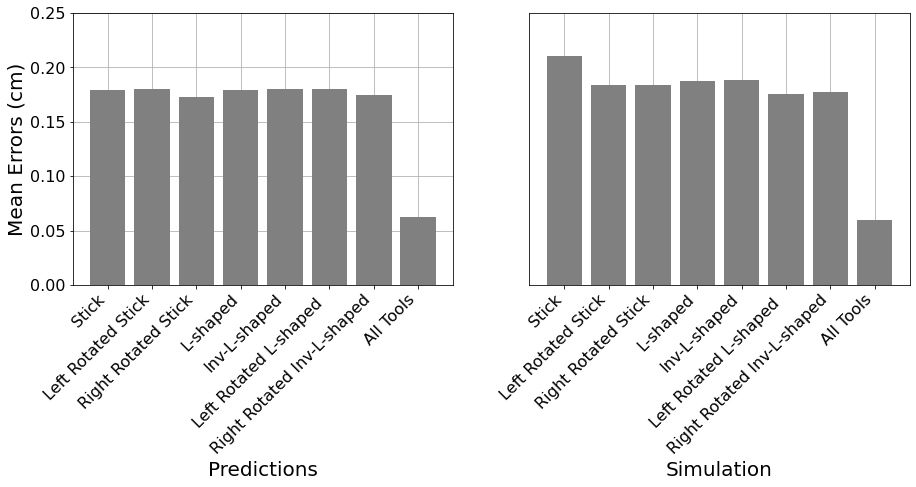}
    \includegraphics[width=0.45\linewidth]{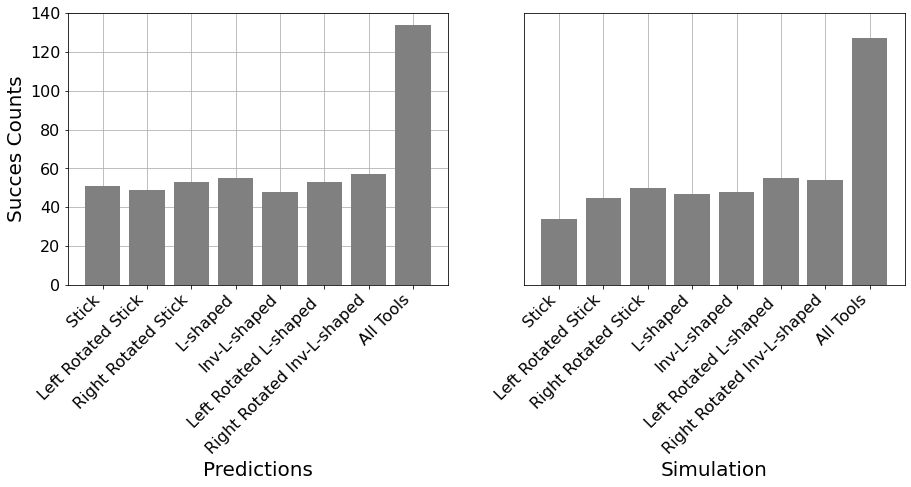}
    \caption{Tool results. As robot is allowed to use wider variety of tools, success rate increases and error amount decreases.}
    \label{fig:tool_results}
\end{figure}

The task is defined as the selection of the best tool and best action sequence from all possible tools and action sequences. The network is run for all the initial-target position pairs for each possible tool and action sequences. For each of these pairs,  the tool and action sequence that gives the lowest mean error is selected. Besides, for comparison, to see whether our framework can utilize each of the tools, the best action sequences for each tool are found as well. Then, each solution is transferred to simulation to testing their correctness.

The results can be seen in Figure~\ref{fig:tool_results}. The left column shows the prediction errors of selected action sequences, and the right column shows actual errors of selected actions when they are run on simulation. The first row shows the mean error between the final positions of manipulated objects and the goal positions. The second row shows the number of successful action sequences (i.e., action sequences where the final position of the object is less than 5 cm away from its target position.). Each bar corresponds to the result for action selection with a particular tool, and with the last one, the tool can be selected as well. From the figure, it can be seen that our framework managed to utilize all tools for solving about 40 of the tasks, and when all tools are allowed to be used, about 130 of the tasks are solvable. Comparing prediction and simulation results shows that predictions made by our framework are plausible, and there is just marginal loss of performance when found action sequences are transferred to simulation. Our framework is successful in tool manipulation and action selection despite its not being designed for such a task.

\begin{figure}[!t]
    \centering

    \includegraphics[trim=3cm 0 1.0cm 1.5cm, clip,width=0.19\linewidth]{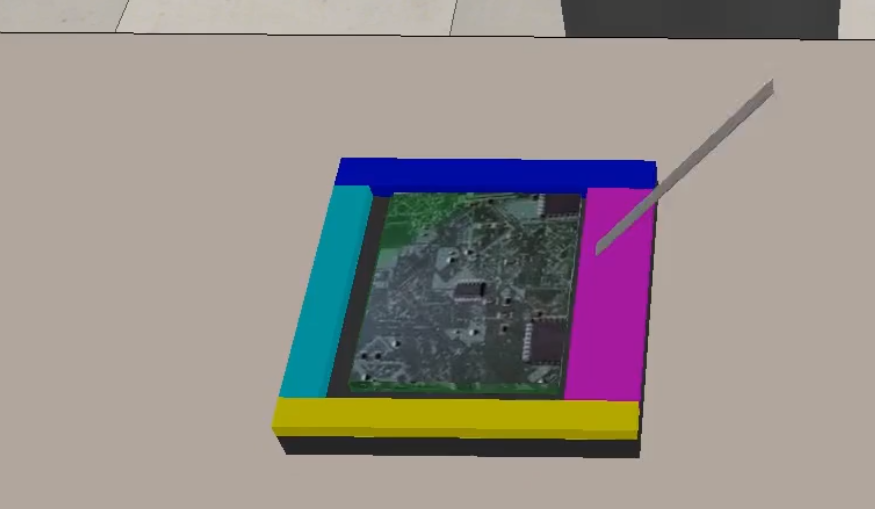}
    \includegraphics[trim=3cm 0 1.0cm 1.5cm, clip,width=0.19\linewidth]{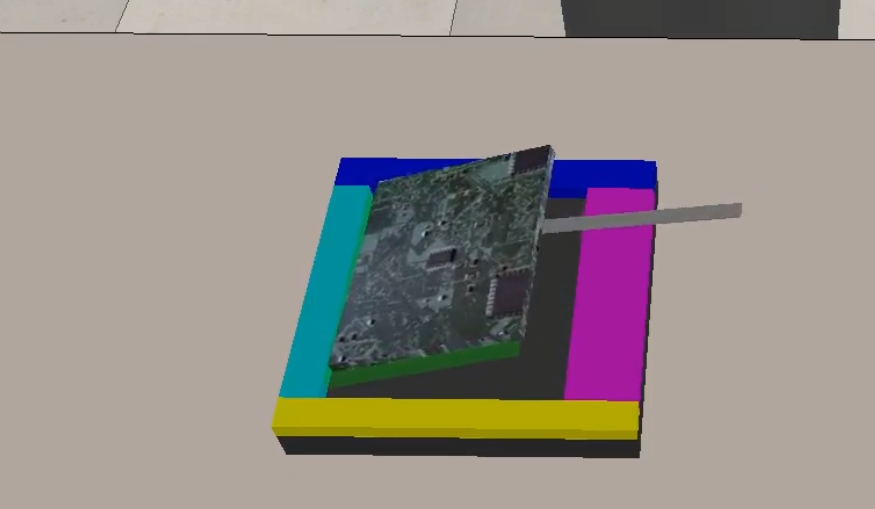}
    \includegraphics[trim=3cm 0 1.0cm 1.5cm, clip,width=0.19\linewidth]{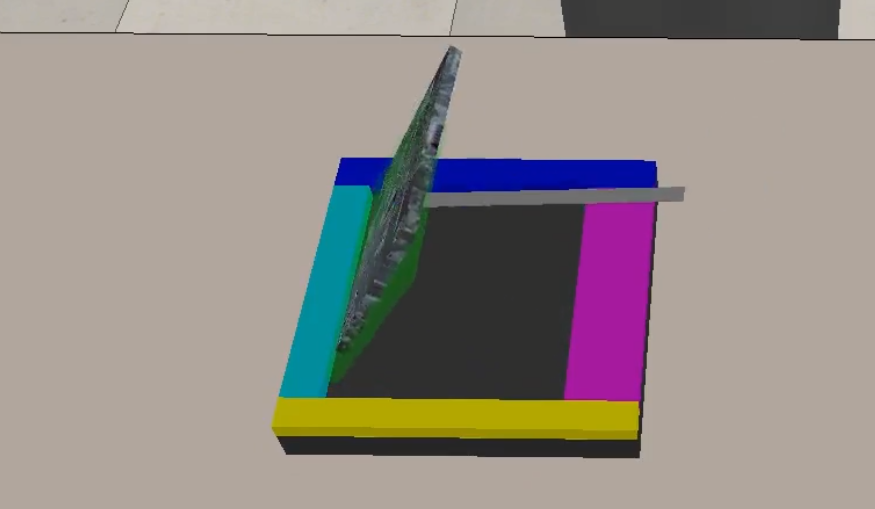}
    \includegraphics[trim=3cm 0 1.0cm 1.5cm, clip,width=0.19\linewidth]{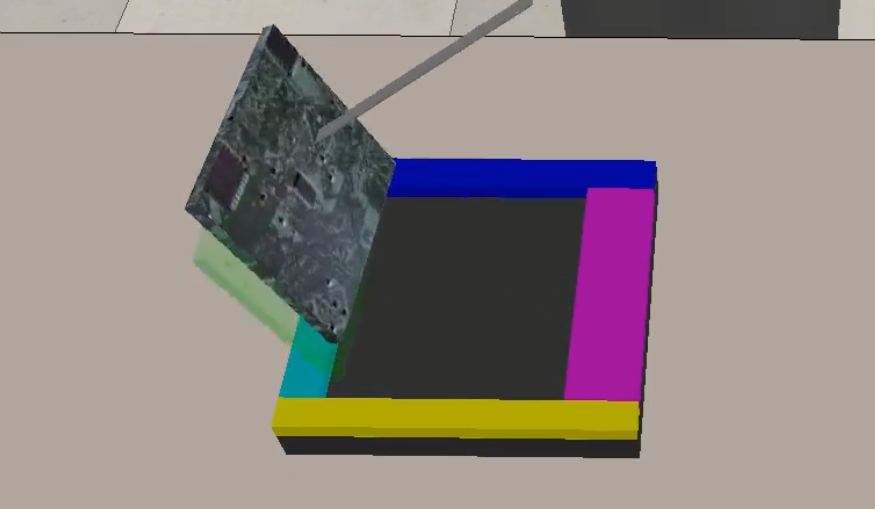}
    \includegraphics[trim=3cm 0 1.0cm 1.5cm, clip,width=0.19\linewidth]{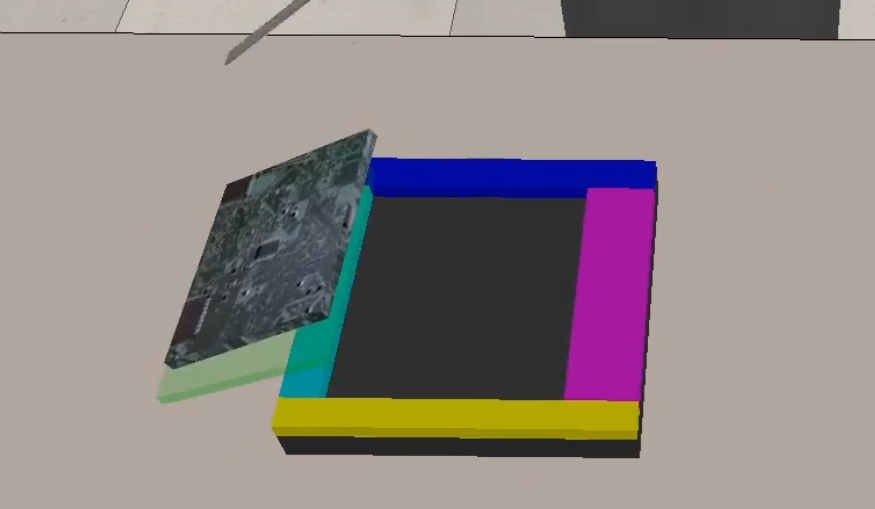}    
    \caption{Snapshots of 6D Effect Prediction. Ground truth pose of object is shown with transparent cuboid. As can be seen, prediction is very close to ground truth.}
    \label{fig:6D_prediction}            
\end{figure}

\subsection{Qualitative Analysis in Simulation - 6D Motion Prediction}

Finally, we designed an experimental setup where we can test our framework on  6D rigid body motion prediction. In this setup, the robot is tasked to lever up a printed circuit board (PCB) from a hard drive disk (HDD) with a screwdriver tool. PCB is on top of the HDD, and at each side of the HDD, there may be a ledge that PCB may contact while being levered up. PCB and HDD are represented as a set of boxes, and their sizes change between runs. Note that some sides of the HDD may have no ledge in different scenes, and therefore, while representing a scene in a graph, the number of nodes changes between runs.

For scene generation, lengths of both sides of HDD are set to $20\ cm$. There is either a ledge of size between $0$ to $8\ cm$, or no ledge at each side of HDD. In the middle of HDD, a PCB with its side lengths between $10$ to $20\ cm$ are generated. 
The network is trained using 500 lever-up interactions on scenes with 125 different procedurally generated hard-disks (One lever-up action from each side of HDD).  

A sample prediction can be seen in Figure~\ref{fig:6D_prediction}. Our further results on this setup can be found on project page. 
In this setup, our network make plausible predictions that match well with the ground truth.

\subsection{Analysis of our framework in real world} \label{sec:real_world}
\begin{figure}[t]
    \centering

    \includegraphics[width=\linewidth]{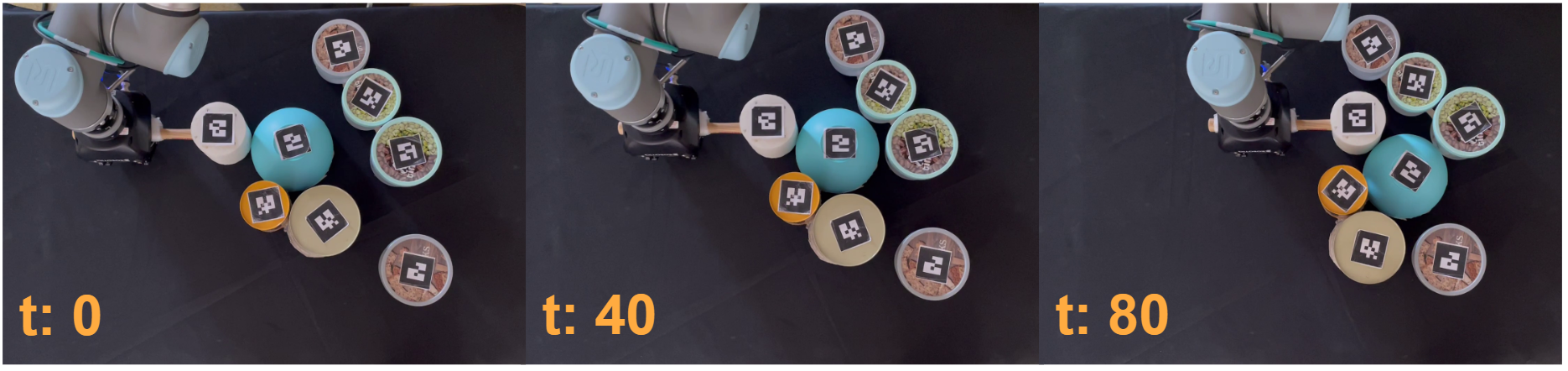}
    
    \caption{Snapshots of a robot interaction in real-world. Our framework continuously updates its joint predictions as it observes the motion of objects and predicts their future positions. }
    \label{fig:real_world_demo}            
\end{figure}

In this section, our framework is evaluated with a real-world dataset, presented in \cite{tekden2019belief}. In this dataset, a UR10 robot arm holds a hammer and use it for pushing objects. The dataset contains cylinder-shaped objects and possible fixed joints between them. The effect of a fixed joint between objects is mimicked by placing customized card-boards under them. A sample created scene and how the robot makes its manipulation on objects can be found in Figure~\ref{fig:real_world_demo}. As the dataset does not have angle information, our network is retrained with the angles of cylinders removed. Since it is also possible for our network to predict revolute or prismatic joint, prediction are limited only to no-joint and fixed joint relations\footnote{Unlike \cite{tekden2019belief}, we do not retrain our network with only cylindrical objects and fixed joints; we only remove angle information of cylindrical objects.} (By selecting the joint relation with the max probability between no-joint and fixed joint relations.).

\begin{figure}[t] 
    \centering
    \includegraphics[width=0.9\linewidth]{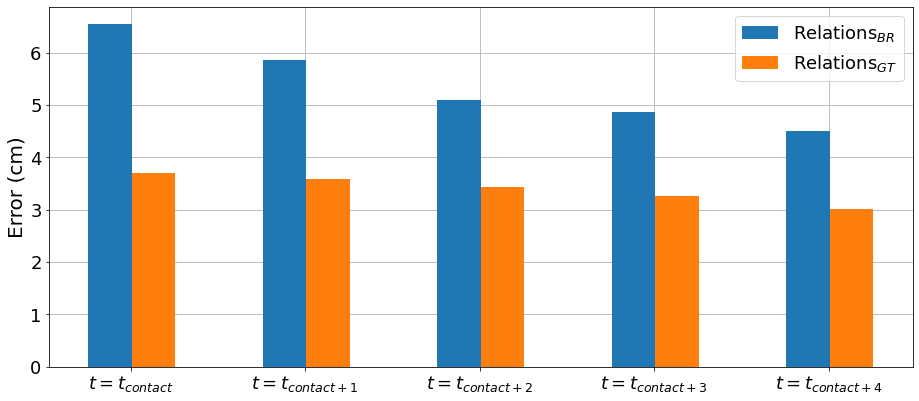}
    
    \caption {Average errors (in cm) change in real world as robot makes its first contact with the objects.}
    \label{fig:MSE-real-contact}
\end{figure}

\begin{figure}[!t]
    \centering
    \includegraphics[width=0.9\linewidth]{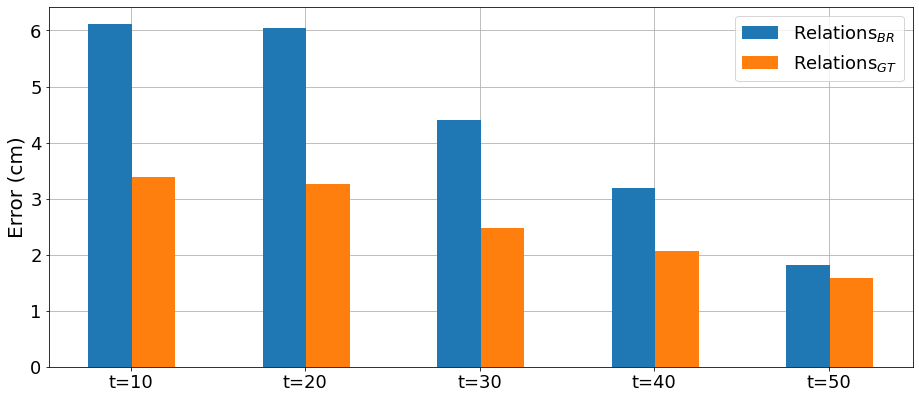}
    
    \caption {Average errors (in cm) change in real-world as our framework acquires more object tracking information.}

    \label{fig:MSE-real-over-time}
\end{figure}

The dataset contains scenes with 2 to 5 cylindrical objects and 1 to 3 fixed joint relations between them. In total, there are 102 different test setups in the dataset. On average, objects move $19.5$ cm, and our physics prediction network achieves $3.5$~cm in predicting final object positions where \cite{tekden2019belief} achieved $6.6$~cm in the same test. Our coupled framework is further analyzed with the same dataset in Figure~\ref{fig:MSE-real-contact}. Similar to \cite{tekden2019belief}, we tested our framework on exact timesteps where the first contact between robot and objects occurs. Our network manages to acquire better results than the one in \cite{tekden2019belief} for both physics prediction with ground truth and with predicted relations (In \cite{tekden2019belief}, prediction with ground truth and predicted relations acquires $6.5\ cm$ and $8.5\ cm$ at time $t$ and $4\ cm$ and $6.5\ cm$ at time $t+4$.). In Figure~\ref{fig:MSE-real-over-time}, performance of our framework on different time-steps is shown where predictions of our framework catch up to the ground truth as more observations are acquired.

%% file: sections/conclusion.tex
\section{Conclusion}

We presented methods and a framework for learning action-effects in object and relation-centric push manipulation tasks. Our framework allows the robot to correct its belief about object and relation parameters as it interacts with the scene and observe the effects of its actions. It then can continuously predict the future dynamics of objects. We have tested belief regulation and physics prediction performance on multiple experiments, including a real-world one. We have shown that our framework can predict joint types in articulated object settings with different object and relation types, masses of objects, and their future motion. We have shown that our framework can be extended for 6D trajectory prediction. Furthermore, we also validated our framework on action selection in a tool manipulation task. Although we do not train a new network that includes situations that are not present in our articulated object setting, our network was successfully transferred to this new domain and succeeded in finding action sequences that complete the given tasks. 

As our framework is very generic, we believe it can be further refined and extended. First, our framework can benefit from intelligent exploration strategies that can generalize to a changing number of objects. In addition, learning of unsupervised representations for objects via interactions can be very powerful for the visual grounding of objects. In future work, we are planning to extend our framework for these adaptations. 

%% file: sections/acknowledgment.tex
\section*{Acknowledgment}
This research has received funding from the European Union’s Horizon
2020 research and innovation programme under grant agreement no. 731761,
IMAGINE; supported by a TUBA GEBIP fellowship awarded to E. Erdem;
and supported by a Tubitak 2210-A scholarship awarded to A.E. Tekden.

The numerical calculations reported in this work were partially performed at TUBITAK ULAKBIM, High Performance and Grid Computing Center (TRUBA resources).